\documentclass[12pt, a4paper]{article}

\usepackage{fontspec}
\usepackage[english]{babel}
\usepackage{amsmath}
\usepackage{amsfonts}
\usepackage{amssymb}
\usepackage{graphicx}
\usepackage{hyperref}
\usepackage{float}
\usepackage{subcaption}
\usepackage{microtype}
\PassOptionsToPackage{hyphens}{url}
\usepackage{hyperref}
\usepackage{microtype}

\newcommand{\AR}[1]{AR_{#1}}

\newcommand{\Rspace}[1]{\mathbb{R}^{#1}}
\newcommand{\XS}{X_S}
\newcommand{\xs}[1]{x^{(#1)}}
\newcommand{\Nsamp}{N_{\text{samples}}}

\newcommand{\Sj}{S_j}

\newcommand{\ARj}{\AR{j}}
\newcommand{\indicate}[1]{\mathbb{I}(#1)}
\newcommand{\vol}{\text{vol}}

\newcommand{\Cuv}{C_{uv}}
\newcommand{\LayerDist}{\text{LayerDistance}}

\newcommand{\ARrestricted}[2]{#1 \big|_{#2}} 

\title{Functional Similarity Metric for Neural Networks: Overcoming Parametric Ambiguity via Activation Region Analysis}
\author{Kutomanov Hennadii} 
\date{\today}

\begin{document}

\maketitle
\tableofcontents
\newpage

\section{Introduction: The Problem of Identification, Comparison and Analysis of Stability of Neural Networks}
\subsection{Relevance}

\begin{itemize}
\item \textbf{Increasing model complexity and the ambiguity problem.}
\end{itemize}
Modern deep neural networks can contain from millions to billions of trainable parameters. As the number of parameters increases, the complexity of the space of possible representations for the same learned function grows exponentially. This fundamental problem of representational ambiguity – when identical or nearly identical functionality can be implemented by numerous different configurations of weight coefficients – becomes increasingly acute. Developing methods to bring models to a canonical form and introducing objective similarity metrics are critically important tasks for in-depth analysis, comparison, and reliable use of neural networks.

\begin{itemize}
\item \textbf{Reproducibility of scientific research.}
\end{itemize}
The lack of standardized approaches for comparing neural networks significantly hampers the reproducibility of results in machine learning. When attempting to reproduce published data, differences in parameter initialization, data presentation order, or the stochastic nature of optimization algorithms inevitably lead to models with different sets of weights, even if the architecture and training hyperparameters are preserved. Without reliable ways to establish functional equivalence or to quantitatively assess the similarity of such models, an objective evaluation of reproduction success becomes impossible, thereby hindering scientific progress.

\begin{itemize}
\item \textbf{Transparency, audit, and regulatory requirements.}
\end{itemize}
The growing deployment of artificial intelligence algorithms in critical areas such as healthcare, finance, autonomous transport, and decision-making systems is accompanied by increased scrutiny from regulatory bodies and society regarding the transparency, reliability, and explainability of models. Effective deployment and safe operation of neural networks in such areas are impossible without objective methods for their characterization, validation, and audit. Canonical representations and functional similarity metrics can greatly simplify certification, verification, and monitoring processes of neural network models, contributing to increased trust in them.

\begin{itemize}
\item \textbf{The proposed approach in the context of existing challenges.}
\end{itemize}
This paper proposes a new approach to addressing the problem of representational ambiguity in neural networks. We introduce the concept of a \textit{canonical form} – an ordering of neurons based on analysis of their functional contribution, characterized through sampled activation regions and represented by efficient hash signatures. Based on this ordering, we develop a \textit{similarity metric} between neural networks of the same architecture. These tools are aimed at creating a foundation for more objective and profound analysis of neural network models.

\subsubsection{Application areas and potential benefits of the proposed approach} 

\begin{itemize}
\item \textbf{Model merging and transfer learning}:
\end{itemize}
When merging neural networks trained on different but potentially overlapping datasets, or for solving related subtasks, a key problem is the correct matching of functionally equivalent or similar components (neurons, layers). The proposed canonical form and similarity metric can provide more accurate alignment of representations, simplifying the matching of functionally identical neurons across networks. This enables more effective knowledge aggregation and prevents loss of specialization when merging models.

\begin{itemize}
\item \textbf{Computational resource optimization and evaluation of model transformations}:
\end{itemize}
Parameter representation optimization techniques, such as weight quantization, pruning, or the use of low-rank approximations, aim to reduce computational costs and memory footprint. The proposed functional similarity metric allows quantitative assessment of how functionally close the optimized (transformed) version of a model is to the original, providing an objective criterion for finding the optimal balance between efficiency and preservation of original functionality without changing the architecture.

\begin{itemize}
\item \textbf{Analyzing learning dynamics and understanding optimization processes}:
\end{itemize}
Tracking functional changes in neural network components throughout the training process is an important task for understanding representation formation mechanisms and the effectiveness of various optimization methods. Due to parameterization ambiguity, it is difficult to separate true functional changes from trivial transitions to equivalent representations. Using canonical forms and a similarity metric enables a more objective evaluation of the model's functional evolution, identification of stable and changing components, and potentially contributes to the development of more advanced training strategies.

\begin{itemize}
\item \textbf{Federated learning}:
\end{itemize}
In federated learning scenarios, where multiple local models are trained on distributed and private data, effective methods are needed to aggregate these models into a single global model. Canonical representation and a similarity metric can facilitate more accurate alignment and averaging of parameters of independently trained models, potentially enhancing the efficiency of federated learning and the quality of the final aggregated model.

\begin{itemize}
\item \textbf{Interpretability, explainability, and compliance with XAI (Explainable AI) requirements}:
\end{itemize}
A standardized and robust representation of neural network components can provide a more reliable foundation for interpretation and explanation methods. Functional hash signatures of neurons, reflecting their characteristic behavior on data, can aid in identifying and cataloging typical functions learned by neurons, simplifying the understanding of model operation principles and increasing its transparency.

\begin{itemize}
\item \textbf{Plagiarism detection and intellectual property protection for models}:
\end{itemize}
With the growing commercial value of trained AI models, protecting intellectual property has become increasingly important. The proposed functional similarity metric can serve as a tool for objectively assessing whether one model is derivative or a copy of another, even if their parameters have been altered (e.g., through fine-tuning, rescaling, or other transformations that do not change the essence of the learned function).

The approach developed in this paper, based on analyzing sampled activation regions of neurons and efficient hashing methods, opens up new opportunities for addressing the aforementioned problems and can contribute to further progress in the analysis, comparison, and reliable application of deep neural networks.

\subsection{The need for a canonical form}

The difficulties noted in the previous section, arising from the representational ambiguity of neural networks, create an urgent need to develop methods for bringing them to a \textbf{canonical form}. In the context of this work, by the canonical form of a neural network (or its individual layer) we mean such a representation that:

\begin{itemize}
    \item \textbf{Is invariant to known symmetries:} It is identical (or maximally similar) for all networks belonging to the same equivalence class, meaning networks that can be obtained from one another via transformations that do not alter their functionality (primarily by scaling neuron parameters and permuting neurons within a layer).
    \item \textbf{Is unique (or tends towards uniqueness):} It allows selection of a single or small number of standard representatives from the entire variety of functionally equivalent parameterizations.
    \item \textbf{Is constructively defined:} It can be obtained via a well-defined algorithm applied to the original network parameters.
\end{itemize}

Thus, the main purpose of introducing a canonical form is to transition from the original, potentially ambiguous parameter space to a certain factor space (or its representation), where each unique functional element has its unambiguous or standardized description.

The existence of such a canonical form is critically important for solving a wide range of tasks:

\begin{itemize}
    \item \textbf{Ensuring objective model comparison:} If two models reduced to canonical form have identical or similar parameters, this suggests their functional closeness.
    \item \textbf{Analyzing structure and learned representations:} Canonical representation can help reveal true structural features of the model, not masked by arbitrary choices of scales or neuron order.
    \item \textbf{Facilitating knowledge transfer and model merging tasks:} Aligning models via their canonical forms can significantly ease the matching and integration of their components .
    \item \textbf{Improving research reproducibility:} Standardizing model representations contributes to more reliable reproduction and verification of scientific results.
\end{itemize}

Developing a practical method for constructing such a canonical form for neural networks with ReLU activations is one of the key objectives of this study.

\subsection{The need for a similarity metric}

While the canonical form discussed in the previous section aims to obtain a unique or standardized representation of a neural network to eliminate ambiguities associated with symmetries , it does not always provide a complete answer when comparing models that are not strictly equivalent. Canonical representation can help establish fundamental similarity or difference at the level of structure and normalized parameters, but often there is a need for a \textbf{quantitative assessment of the degree of this similarity or difference}.

It is for this purpose that the concept of a \textbf{similarity metric (or measure)} between neural networks (or their corresponding layers) of the same architecture is introduced. If the canonical form answers the question: "Are these two models essentially the same?", the metric answers the question: "How close or far are they from each other if they are not identical?" 

The need for such a quantitative measure is driven by the fact that:

\begin{itemize}
    \item It allows \textbf{ranking} models by their degree of closeness to a reference or to each other, which is important for selection or analysis .
    \item It makes it possible to \textbf{track continuous changes} in a model (e.g., during training \mbox{[CITE: works on learning dynamics analysis]} or when adapting to new data), not just discrete transitions between equivalence classes.
    \item It provides a basis for \textbf{establishing threshold values} for decision-making (e.g., whether two models are "similar enough" for a particular purpose \mbox{[CITE: examples where similarity thresholds are used]}).
\end{itemize}

Thus, developing an adequate similarity metric is a logical extension and complement to the concept of canonical representation, providing a more fine-grained and flexible tool for neural network analysis. This study aims not only to propose a method for obtaining standardized neuron characteristics but also, based on them, to define a computationally efficient and interpretable similarity metric.

\subsection{Goals and objectives of the paper}

Based on the relevance and existing needs in neural network analysis described above, the \textbf{main goal} of this work is to develop a new method for obtaining a standardized representation of neural layers and, based on it, a quantitative metric of their functional similarity. This will enable more objective comparison and analysis of neural networks with identical architectures.

To achieve this goal, the following \textbf{specific tasks} are addressed in this study:
\begin{enumerate}
    \item Development of an algorithm to obtain characteristic signatures of individual neurons based on the analysis of their behavior, using efficient hashing methods to compactly represent these characteristics.
    \item Creation of a procedure to establish optimal pairwise matching between neurons of two compared layers.
    \item Formal definition and introduction of a similarity (or distance) metric between neural layers, quantitatively reflecting the degree of their functional closeness based on the established matches.
    \item Theoretical analysis of the developed algorithm, including assessment of its computational complexity and investigation of the mathematical properties of the proposed metric.
    \item Experimental evaluation of the proposed approach to assess its robustness, sensitivity, and practical applicability in tasks of neural network comparison and analysis.
\end{enumerate}

\subsection{Central novelty of the proposed approach}

The central novelty of this work lies in shifting the problem of neural network comparison from the parameter space to the space of functional characteristics. For ReLU networks, the same function can be realized by different sets of weights due to scaling and neuron permutation symmetries; therefore, direct parameter comparison does not yield a reliable measure of functional similarity. 

We propose an approach based on analyzing neuron activation regions and their sampled data signatures, which enables the construction of a \textbf{canonical layer representation} and the definition of a computationally feasible measure of functional similarity between layers and networks of identical architecture.

For both the continuous and discrete settings, a metric foundation is established via the \textbf{Jaccard distance}, and for the practical \textbf{MinHash}-based implementation, theoretical guarantees on approximation accuracy and distance ordering preservation are
provided. This makes the proposed method simultaneously theoretically grounded and applicable to real-world networks.

\subsection{Structure of the paper}

This paper is organized as follows.

\textbf{Section 2} lays the theoretical foundation: it considers the main symmetries inherent in neural networks with ReLU activations and discusses the procedure of L2 normalization of neuron parameters as an initial step towards eliminating representational ambiguity.

\textbf{Section 3} reviews existing approaches to neuron ordering or labeling, analyzing their advantages and key shortcomings, particularly the problem of instability to small parameter changes.

\textbf{Section 4} introduces the conceptual notion of a neuron's "activation region" ($AR_j$) in the continuous input space, discussing its geometric properties and theoretical significance for characterizing the neuron's function.

\textbf{Section 5} transitions from the theoretical model of the continuous activation region to its practical characterization on discrete data. It introduces the notion of the "activation region of neuron $j$ restricted to a sample $X_S$" (denoted $AR_j|_{X_S}$). Methodological aspects of forming the sample $X_S$ are discussed.

\textbf{Section 6} describes methods for analyzing the obtained restricted activation regions $AR_j|_{X_S}$ (and their binary vector representations), including extraction of primary neuron characteristics and assessment of their pairwise similarity.

\textbf{Section 7} is devoted to methods for efficiently representing $AR_j|_{X_S}$ characteristics using hashing techniques such as Bloom Filter and MinHash. Special attention is given to their ability to preserve similarity information while significantly compressing data.

\textbf{Section 8} presents a detailed description of the developed algorithm for establishing correspondence between neurons of two layers based on hashed representations of their $AR_j|_{X_S}$. This section concludes with the formal definition of a similarity metric between layers, computed based on the results of this matching.

\textbf{Section 9} analyzes the algorithmic complexity of the approach proposed in Section 8, including complexity estimates for each stage.

\textbf{Section 10} is dedicated to the theoretical analysis of the properties of the introduced similarity metric, including its formal mathematical characteristics (e.g. non-negativity, symmetry, satisfaction or violation of the triangle inequality).

\textbf{Section 11} presents the results of experimental evaluation of the proposed method and metric. Their robustness to small parameter perturbations, sensitivity to functional model differences, and applicability to practical neural network comparison and analysis tasks are investigated.

In \textbf{Section 12 (Conclusion)}, the main results of the work are summarized, its limitations are discussed, and promising directions for future research are outlined.

\section{Theoretical basis: symmetries in ReLU networks and preliminary parameter normalization}
\label{sec:theoretical_basis}
\subsection{Scaling \texorpdfstring{($D$)}{(D)} and permutation \texorpdfstring{($P$)}{(P)} transformation groups preserving ReLU network functions}

The parameter space $\Theta$ of a neural network $\mathcal{N}$ (with fixed architecture) admits a number of transformations that do not change the function $f_{\mathcal{N}}: \mathbb{R}^{N_{in}} \to \mathbb{R}^{N_{out}}$ realized by the network. These transformations form groups, and understanding them is key to the problem of representational ambiguity. For fully connected networks with ReLU (Rectified Linear Unit) activation functions, two main groups of such transformations can be distinguished: the group of positive diagonal scalings and the group of permutations.

\subsubsection{Positive diagonal scaling group \texorpdfstring{($D$)}{(D)}}

Consider neuron $j$ in layer $k$ of a network. Its parameters are the vector of incoming weights $W_j^{(k)}$ and the bias $b_j^{(k)}$. Its output $a_j^{(k)} = \text{ReLU}(W_j^{(k)} x^{(k-1)} + b_j^{(k)})$ is passed to the inputs of neurons in the next layer $k+1$ via the outgoing weights $W_{\cdot, j}^{(k+1)}$ (we assume $j$ indexes the columns of matrix $W^{(k+1)}$).

Due to the positive homogeneity property of ReLU ($\text{ReLU}(c z) = c \cdot \text{ReLU}(z)$ for $c > 0$), we can apply a scaling transformation to each neuron $j$ in layer $k$. Let $D^{(k)}$ be a diagonal matrix of size $M_k \times M_k$ (where $M_k$ is the number of neurons in layer $k$) with strictly positive diagonal entries $D_{jj}^{(k)} = c_j > 0$. The action of $D^{(k)}$ on the network parameters is defined as:

\begin{enumerate}
    \item \textbf{Incoming parameters of layer $k$ neurons:} If $W^{(k)}$ is the weight matrix of layer $k$ (rows correspond to neurons) and $b^{(k)}$ is the bias vector, then:
    $$(W^{(k)}, b^{(k)}) \xrightarrow{D^{(k)}} (D^{(k)}W^{(k)}, D^{(k)}b^{(k)})$$
    (Each $j$-th row $W_j^{(k)}$ and $j$-th element $b_j^{(k)}$ are multiplied by $c_j$).
    \item \textbf{Outgoing weights from layer $k$ neurons (i.e. columns of matrix $W^{(k+1)}$):}
    $$W^{(k+1)} \xrightarrow{D^{(k)}} W^{(k+1)}(D^{(k)})^{-1}$$
    (Each $j$-th column $W_{\cdot,j}^{(k+1)}$ is divided by $c_j$, as $(D^{(k)})^{-1}$ is diagonal with entries $1/c_j$).
\end{enumerate}

Under this transformation, the pre-activation vector of layer $k$, $z'^{(k)} = D^{(k)}z^{(k)}$, and the activation vector $a'^{(k)} = D^{(k)}a^{(k)}$. The contribution to the pre-activations of the next layer remains unchanged: $(W^{(k+1)}(D^{(k)})^{-1})(D^{(k)}a^{(k)}) = W^{(k+1)}a^{(k)}$. Thus, the network function $f_{\mathcal{N}}$ is invariant.

The set of such diagonal matrices $D^{(k)}$ with $D_{jj}^{(k)} > 0$ forms a group under matrix multiplication, isomorphic to $(\mathbb{R}_{>0})^{M_k}$.

\subsubsection{Neuron permutation group \texorpdfstring{($P$)}{(P)}}
\label{subsec:permutation_group} 

Consider hidden layer $k$ containing $M_k$ neurons. Let $P^{(k)}$ be an arbitrary permutation matrix of size $M_k \times M_k$ (each row and column has one entry of one and the rest are zeros), corresponding to a permutation $\pi \in S_{M_k}$. The action of $P^{(k)}$ on the network parameters is defined as:

\begin{enumerate}
    \item \textbf{Incoming parameters of layer $k$ neurons:}
    $$(W^{(k)}, b^{(k)}) \xrightarrow{P^{(k)}} (P^{(k)}W^{(k)}, P^{(k)}b^{(k)})$$
    (Rows of $W^{(k)}$ and entries of $b^{(k)}$ are permuted according to $P^{(k)}$).
    \item \textbf{Outgoing weights from layer $k$ neurons (i.e. columns of matrix $W^{(k+1)}$):}
    $$W^{(k+1)} \xrightarrow{P^{(k)}} W^{(k+1)}(P^{(k)})^T$$
    (Columns of $W^{(k+1)}$ are permuted accordingly, since for permutation matrices $(P^{(k)})^{-1} = (P^{(k)})^T$).
\end{enumerate}

Such simultaneous application of $P^{(k)}$ to layer $k$ neuron indices when accessing their incoming parameters (effectively permuting rows of $W^{(k)}$ and $b^{(k)}$) and their outputs when forming the inputs of the next layer (effectively permuting columns of $W^{(k+1)}$) guarantees that the function computed by $f_{\mathcal{N}}$ remains invariant. For each hidden layer $k$ with $M_k$ neurons, there are $M_k!$ such permutation matrices forming a discrete group isomorphic to the symmetric group $S_{M_k}$.

These two transformation groups – the positive diagonal scaling group (represented by $D$ matrices) and the permutation group (represented by $P$ matrices) – are the main sources of parameter representation ambiguity in fully connected ReLU networks. In the next subsection, we consider how they naturally combine into the group of positive monomial matrices $S=DP$.

\subsection{Positive monomial transformations, intertwining condition, and interaction with ReLU}

The transformation groups considered in the previous subsection – positive diagonal scalings (represented by matrices $D^{(k)}$) and permutations (represented by matrices $P^{(k)}$) – are fundamental symmetries preserving the function of ReLU networks. These two types of transformations naturally combine into a more general algebraic structure: the \textbf{group of positive monomial matrices}.

\subsubsection{Positive monomial matrices \texorpdfstring{($S=DP$)}{(S=DP)}}

\textbf{Definition:} A positive monomial matrix $S$ of size $M \times M$ is a matrix in which each row and each column contains exactly one non-zero entry, and this entry is strictly positive. Any such matrix can be represented as a product:
$$S = DP$$
where:
\begin{itemize}
    \item $D$ is a diagonal matrix of size $M \times M$ with strictly positive diagonal elements ($D_{ii} > 0$). This matrix is responsible for individually scaling neuron outputs.
    \item $P$ is a permutation matrix of size $M \times M$. This matrix is responsible for reordering neuron outputs.
\end{itemize}
The set of all $M \times M$ positive monomial matrices forms a group under matrix multiplication, denoted $G_{pm}(M)$ \cite{tran2024monomial}. This group includes as special cases the group of positive diagonal scalings (when $P=I$, the identity matrix) and the permutation group (when $D=I$).

\subsubsection{“ReLU-compatibility” and transformation transfer}

A key property of positive monomial matrices is their "compatibility" with the ReLU activation function. As established in \cite{tran2024monomial}, the equality:
$$\text{ReLU}(SZ) = S \cdot \text{ReLU}(Z)$$
holds for all vectors $Z \in \mathbb{R}^M$ if and only if $S$ is a positive monomial matrix. This property allows the transformation $S$ to be "factored out" from the ReLU function, which is fundamental for analyzing equivalent transformations in the network.

Consider a scenario where some effective transformation acts on the input to the linear part of a layer (before activation). Suppose the argument of the ReLU function in the first layer is $(W_1R)x + b_1$, where $R$ is a matrix representing some input linear transformation. We want to understand when the effect of $R$ can be equivalently transferred \textit{after} applying ReLU in the form of a matrix $S_{\text{out}}$ acting on the ReLU outputs:
$$\text{ReLU}((W_1R)x + b_1) = S_{\text{out}} \cdot \text{ReLU}(W_1x + b_1'')$$
For this, there must exist a relationship between the ReLU arguments. One way to establish such equivalence is via the \textbf{algebraic intertwining condition}. If we can find a positive monomial matrix $S_{\text{in}}$ such that the arguments are related by $(W_1R)x + b_1 = S_{\text{in}}(W_1x + b_1'')$, then due to the “ReLU-compatibility” of $S_{\text{in}}$, we obtain:
$$\text{ReLU}(S_{\text{in}}(W_1x + b_1'')) = S_{\text{in}} \cdot \text{ReLU}(W_1x + b_1'')$$
In this case, $S_{\text{out}} = S_{\text{in}}$.

\subsubsection{Algebraic intertwining condition \texorpdfstring{$W_1R = SW_1$}{W1R = SW1}}

The condition $(W_1R)x + b_1 = S(W_1x + b_1'')$ with $S=S_{\text{in}}$ leads to two equalities when equating coefficients of $x$ and constant terms:
\begin{enumerate}
    \item $W_1R = SW_1$
    \item $b_1 = Sb_1'' \implies b_1'' = S^{-1}b_1$ (since $S$, as a positive monomial matrix, is invertible).
\end{enumerate}
The equation $W_1R = SW_1$ is central. Here, $W_1$ is the weight matrix of the first layer, $R$ is the input transformation matrix, and $S$ is a positive monomial matrix characterizing the transformation in the pre-activation space of this layer. This condition means that applying $R$ to the input and then $W_1$ is equivalent to applying $W_1$ first and then $S$ in the pre-activation space.

\begin{itemize}
    \item If $W_1$ is invertible (rare in typical layers), then $R = W_1^{-1}SW_1$, i.e., $R$ and $S$ are similar \mbox{[CITE: standard linear algebra]}.
    \item In the general case, this condition imposes significant structural constraints on $W_1$ and $R$, requiring their actions to be consistent so that $W_1R$ can be represented as $SW_1$ for some positive monomial $S$.
\end{itemize}

\subsubsection{Structure \texorpdfstring{$R = S_{\text{in}} + K$}{R = Sin + K} and the role of the null space of \texorpdfstring{$W_1$}{W1}}

Not every arbitrary input transformation $R$ will satisfy the condition $W_1R = SW_1$ for some “ReLU-compatible” $S$. However, the effect of $R$ on the output of the layer’s linear part $W_1Rx$ depends only on the part of $R$ not annihilated by $W_1$.

Consider the decomposition $R = S_{\text{in}} + K$, where:
\begin{itemize}
    \item $S_{\text{in}}$ is the component of $R$ for which we seek a corresponding positive monomial $S$ such that $W_1S_{\text{in}} = SW_1$.
    \item $K$ is the component of $R$ such that $W_1K = \mathbf{0}$ (zero matrix). This means that for any input $x$, the vector $Kx$ belongs to the null space (kernel) of $W_1$.
\end{itemize}
If such a decomposition exists, then the linear operation of the first layer yields:
$$W_1(Rx) = W_1((S_{\text{in}}+K)x) = W_1S_{\text{in}}x + W_1Kx = W_1S_{\text{in}}x + \mathbf{0} = W_1S_{\text{in}}x$$
Thus, the component $K$ is “filtered out” or “absorbed” by the first layer and does not affect the activation argument. Only the component $S_{\text{in}}$ (or more precisely, its product $W_1S_{\text{in}}$) determines the pre-activations. If for this $S_{\text{in}}$ the condition $W_1S_{\text{in}} = SW_1$ with some positive monomial $S$ holds, then the effect of $S_{\text{in}}$ can be “transferred” through ReLU as $S$.

This means that for such “transfer” what matters is not $R$ itself, but its part that is “visible” to $W_1$ and compatible with some positive monomial $S$ via the intertwining condition.

\subsection{Preliminary L2 normalization of parameters as a way to fix the scaling component}
\label{subsec:l2_normalization}

The first practical step towards a canonical representation of a neural network, aimed at eliminating the ambiguity introduced by the positive diagonal scaling group $D$ (see Subsection 2.1.1), is the procedure of L2 normalization of each layer’s parameters. This procedure is performed sequentially, layer by layer, starting from the first hidden layer. The main goal of normalization is to bring neuron weight vectors to a common $L_2$ norm, making their orientations directly comparable and stabilizing subsequent analysis stages. Simultaneously, adjustments are made to the weights of the next layer to compensate for these changes, thus preserving the functional equivalence of the network.

\subsubsection{Notation for layer \texorpdfstring{$k$}{k}}

Consider an arbitrary layer $k$ of a neural network.
\begin{itemize}
    \item $N_k$ – the number of neurons in layer $k$.
    \item $D_{in}^{(k)}$ – the input vector dimension for layer $k$.
    \item $W_{k,j} \in \mathbb{R}^{D_{in}^{(k)}}$ – the weight vector of neuron $j$ ($1 \le j \le N_k$) in layer $k$.
    \item $b_{k,j} \in \mathbb{R}$ – the bias of neuron $j$ in layer $k$.
    \item $x \in \mathbb{R}^{D_{in}^{(k)}}$ – the input vector for layer $k$.
    \item $z_{k,j}(x) = W_{k,j} \cdot x + b_{k,j}$ – the pre-activation (logit) of neuron $j$ in layer $k$ for input $x$.
\end{itemize}

\subsubsection{Parameter normalization procedure for layer \texorpdfstring{$k$}{k}}

For each neuron $j$ in layer $k$, its parameters $(W_{k,j}, b_{k,j})$ are transformed into normalized parameters $(W'_{k,j}, b'_{k,j})$ and a scaling factor $c_{k,j}$ is computed as follows:

\paragraph{Calculating the $L_2$ norm of the weight vector.}
Compute the Euclidean norm ($L_2$ norm) of the neuron's weight vector:
\[ \rho_{k,j} = ||W_{k,j}||_2 = \sqrt{\sum_{i=1}^{D_{in}^{(k)}} (W_{k,j,i})^2} \]

\paragraph{Normalizing parameters based on the norm value.}
\begin{itemize}
    \item \textbf{Case 1: $\rho_{k,j} \neq 0$} (standard case). \\
    Here, the neuron defines a hyperplane with a well-defined orientation.
    \begin{itemize}
        \item Scaling factor: $c_{k,j} = \rho_{k,j}$.
        \item Normalized weight vector: $W'_{k,j} = \frac{W_{k,j}}{c_{k,j}}$, ensuring $||W'_{k,j}||_2 = 1$.
        \item Normalized bias: $b'_{k,j} = \frac{b_{k,j}}{c_{k,j}}$.
    \end{itemize}

    \item \textbf{Case 2: $\rho_{k,j} = 0$} (i.e. $W_{k,j} = \vec{0}$). \\
    Here, the weight vector is zero. Its output $z_{k,j}(x) = b_{k,j}$ is a constant.
    \begin{itemize}
        \item Normalized weight vector: $W'_{k,j} = W_{k,j} = \vec{0}$.
        \item Normalized bias: $b'_{k,j} = b_{k,j}$.
        \item Scaling factor: $c_{k,j} = 1$.
    \end{itemize}
    (Note: for such neurons, $||W'_{k,j}||_2 = 1$ does not hold. These neurons do not define hyperplanes and may require special treatment in subsequent analysis.)
\end{itemize}

After normalization, the pre-activation of the neuron with new parameters $W'_{k,j}$ and $b'_{k,j}$ is $z'_{k,j}(x) = W'_{k,j} \cdot x + b'_{k,j}$. The relationship between the original and new pre-activation is: $z_{k,j}(x) = c_{k,j} z'_{k,j}(x)$.

\subsubsection{Compensating scaling factors in the next layer \texorpdfstring{($k+1$)}{(k+1)}}

To preserve the overall function of the network, the scaling factors $\{c_{k,j}\}_{j=1}^{N_k}$ of layer $k$ are used to adjust the weights of the next layer $k+1$. If the activation function $\phi$ is positively homogeneous of degree one (e.g. ReLU), then $a_{k,j} = c_{k,j} a'_{k,j}$. Then the weights of layer $k+1$ are modified:
\[ W'_{k+1,p,j} = W_{k+1,p,j} \cdot c_{k,j} \]
for all neurons $p$ in layer $k+1$ and all neurons $j$ in layer $k$. In matrix form:
\[ W'_{k+1} = W_{k+1} \cdot \operatorname{diag}(c_{k,1}, c_{k,2}, \dots, c_{k,N_k}) \]
The biases $b_{k+1}$ of layer $k+1$ remain unchanged.

\subsubsection{Normalization result for layer \texorpdfstring{$k$}{k}}

Upon completing normalization for layer $k$:
\begin{itemize}
    \item Neuron parameters in layer $k$ are replaced with $(W'_{k,j}, b'_{k,j})$.
    \item Weights of layer $k+1$ are modified to $W'_{k+1}$.
    \item The network (up to the output of layer $k+1$) remains functionally equivalent to the original (assuming homogeneous activation).
\end{itemize}

\subsubsection{Processing the output layer}

The described $L_2$ normalization procedure with scaling factor compensation in the next layer is applied to all \textbf{hidden layers} of the neural network. However, the \textbf{last (output) layer ($L$)} is handled specially:

\textbf{No normalization of output layer parameters is performed.} The weights $W_L$ and biases $b_L$ of the output layer neurons are not subjected to the $L_2$ normalization procedure described in Section 1.3. This is because:
\begin{itemize}
    \item There is no subsequent layer to compensate for potential scaling factors $c_{L,j}$ that would arise when normalizing layer $L$. Any scaling of the output layer weights directly changes the scale of the network outputs and thus its function.
    \item Absolute values of the outputs of the final layer (e.g. logits for classification or predicted values for regression) are usually critically important and should not be changed arbitrarily.
    \item The activation function in the output layer (if present, such as Softmax or Sigmoid) often does not have the positive homogeneity property required for correct compensation.
\end{itemize}

\textbf{The output layer weights $W_L$ accumulate compensation from the penultimate layer.} The scaling factors $\{c_{L-1,j}\}_{j=1}^{N_{L-1}}$ obtained during normalization of the penultimate hidden layer $L-1$ are used to adjust (compensate) the output layer weights $W_L$. As described in Section 1.4, the weight matrix $W_L$ is transformed into $W'_L$:
\[ W'_{L} = W_{L} \cdot \operatorname{diag}(c_{L-1,1}, c_{L-1,2}, \dots, c_{L-1,N_{L-1}}) \]
The output layer biases $b_L$ remain unchanged during this compensation.

Thus, as a result of the normalization process, the parameters of all hidden layers are brought to a common $L_2$ norm for their weight vectors, and the output layer weight matrix $W_L$ is adjusted to account for the scaling applied in the last hidden layer. The output layer parameters $(W'_L, b_L)$ are not further normalized.

\subsection{Geometric interpretation of L2-normalized parameters}
\label{subsec:geometry_interpretation_l2}

The L2 normalization procedure applied to hidden layer neuron parameters not only eliminates the ambiguity associated with arbitrary scaling (the $D$ component of the positive monomial transformation group) but also brings the equation of each separating hyperplane $W'_{k,j} \cdot x + b'_{k,j} = 0$ to its \textbf{canonical normal form}. In this form, neuron parameters acquire a clear geometric meaning:
\begin{itemize}
    \item The normalized weight vector $W'_{k,j}$ (obtained as $W_{k,j} / ||W_{k,j}||_2$) represents a \textbf{unit normal vector} to the hyperplane. It uniquely defines the orientation of the hyperplane in the input space $\mathbb{R}^{D_{in}^{(k)}}$.
    \item The normalized bias $b'_{k,j}$ (obtained as $b_{k,j} / ||W_{k,j}||_2$) is interpreted as the \textbf{signed distance from the origin to the hyperplane} along the direction defined by the normal $W'_{k,j}$. The sign of $b'_{k,j}$ indicates on which side of the hyperplane the origin lies (e.g., if $W'_{k,j} \cdot x + b'_{k,j} > 0$ defines the “positive” half-space, then for $b'_{k,j} > 0$ the origin lies in the “negative” half-space, and for $b'_{k,j} < 0$ – in the “positive” half-space). The absolute value $|b'_{k,j}|$ equals this shortest distance.
\end{itemize}
Thus, after L2 normalization, each hidden layer neuron (except those with an original zero weight vector) is characterized by a unit normal vector and a signed distance, providing a standardized and geometrically interpretable representation.

\section{Traditional approaches to neuron ordering and alignment}
\label{subsec:traditional_ordering_alignment_methods}

To overcome issues related to permutation ambiguity (discussed in Subsection~\ref{subsec:permutation_group}) and to enable meaningful comparison or aggregation of neural networks, various methods have been proposed. These methods aim either to establish some order among neurons within a single layer or, more commonly and relevant to practical tasks, to perform \textbf{alignment} of neurons between two or more networks. Below we review the main groups of such approaches applied to pre-L2-normalized parameters (as described in Subsection~\ref{subsec:l2_normalization}). For each method, potential limitations will also be indicated, primarily related to instability.

\subsection{Parameter-based methods}
\label{sssec:parameter_based_methods}

The simplest heuristics for ordering or matching neurons rely directly on their weight and bias values.

\paragraph{Lexicographic sorting.}
This approach involves ordering neurons based on lexicographic comparison of their parameter vectors \cite{parkinson2023relu}.
\begin{itemize}
    \item \textbf{Criteria:} Sorting can be performed based on the incoming normalized weights $W'_{k,j}$, the outgoing weights $W'^{(k+1)}$ (adjusted after normalization of layer $k$), or a concatenation of various parameters ($W'_{k,j}$, $b'_{k,j}$, etc.).
    \item \textbf{Justification:} It is assumed that neurons with similar functions may have lexicographically orderable parameter patterns.
    \item \textbf{Drawbacks and instability:} The main drawback is high sensitivity to small changes in weight values. If the components of two neurons' vectors are very close, the slightest perturbation can change their relative order in lexicographic sorting, making such an order unstable and unreliable for comparison.
\end{itemize}

\paragraph{Sorting based on normalized bias values.}
Neurons are ordered by the scalar values of their normalized biases $b'_{k,j}$ \cite{parkinson2023relu}.
\begin{itemize}
    \item \textbf{Criterion:} The values $b'_{k,j}$ (the signed distance from the origin to the neuron's hyperplane).
    \item \textbf{Justification:} It is assumed that neurons with similar $b'_{k,j}$ may play similar roles. Biases are important parameters in precise ReLU network models, such as MILP formulations \cite{ataei2025mathematical}.
    \item \textbf{Drawbacks and instability:} Similar to lexicographic weight sorting, if bias values of different neurons are close, small changes can easily switch their order. This method also does not account for hyperplane orientation.
\end{itemize}

\subsection{Methods using neuron activation statistics}
\label{sssec:activation_stats_methods}

These approaches order or match neurons based on statistical properties of their activations computed over a representative dataset $\XS$.

\begin{itemize}
    \item \textbf{Criteria:}
        \begin{itemize}
            \item \textbf{Mean activation:} Sorting by mean activation value $\mathbb{E}[a'_{k,j}]$ on $\XS$.
            \item \textbf{Activation variance:} Sorting by variance $\text{Var}[a'_{k,j}]$ on $\XS$.
            \item \textbf{Activation frequency (sparsity):} Ordering by the fraction of input samples from $\XS$ for which the neuron is active. Works such as \cite{lyle2024normalization} discuss metrics like “average dormancy” that can be used here.
        \end{itemize}
    \item \textbf{Justification:} It is assumed that neurons with similar activation profiles on data are functionally similar. This approach takes into account the dynamic behavior of neurons. Model merging works \cite{tatro2020optimizing, li2016convergent} successfully use activation correlation to align neurons between networks.
    \item \textbf{Drawbacks and instability:} Activation statistics can fluctuate with slight weight changes, especially for inputs near neuron decision boundaries. Results can also depend on the specific dataset $\XS$ used and its size. Activation patterns may “flicker,” making derived statistics unstable.
\end{itemize}

\subsection{More advanced algorithmic approaches (often for network alignment)}
\label{sssec:advanced_alignment_methods}

There exist more complex methods primarily aimed at aligning neurons between two networks, closely related to our task of constructing a similarity metric.
\begin{itemize}
    \item \textbf{Optimal assignment-based methods:} “Git Re-Basin: Merging Models modulo Permutation Symmetries” \cite{ainsworth2022git} and “Optimizing Mode Connectivity via Neuron Alignment” \cite{tatro2020optimizing} solve a linear assignment problem to match neurons based on weight similarity or activation correlation, finding an optimal permutation for alignment.
    \item \textbf{Combinatorial structure analysis (sign patterns):} Studies such as \cite{masden2022algorithmic} examine the combinatorial structure of linear regions (chambers) defined by ReLU networks via sign patterns of pre-activations. In theory, characteristics of these patterns or their associated regions could be used for canonical representation, but it is computationally expensive.
    \item \textbf{Stable neuron identification and pruning:} Methods that identify consistently active/inactive neurons \cite{serra2021scaling} or perform pruning based on weight magnitudes implicitly rank neurons by importance, which can be used for ordering.
\end{itemize}

\paragraph{General note.} Most advanced methods in the literature focus on the problem of aligning two networks for merging or mode connectivity analysis rather than defining an autonomous, stable canonical form for a single network based on simple heuristics. This highlights the complexity of the latter task, primarily due to instability issues, which are discussed in detail in the next subsection.

\begin{table}[H]
\centering
\caption{Summary of traditional heuristic neuron ordering/alignment methods and their main limitations.}
\label{tab:traditional_ordering_summary_revised}
\begin{tabular}{|p{4cm}|p{4.5cm}|p{6.5cm}|}
\hline
\textbf{Method name} & \textbf{Ordering/alignment criterion} & \textbf{Justification/Assumption and main drawbacks (instability)} \\
\hline
Lexicographic sorting (by incoming weights/biases/combined) & Normalized vectors $W'_{in}$, $b'$, or their combinations & Assumption: functionally similar neurons have lexicographically orderable parameters. \newline \textbf{Drawback:} Highly sensitive to small parameter changes when values are close. \\
\hline
Sorting by normalized bias values & Scalar value of normalized bias $b'$ & Assumption: neurons with similar $b'$ (similar distance to origin) may play similar roles. \newline \textbf{Drawback:} Highly sensitive if bias values are close; does not account for orientation. \\
\hline
Sorting by activation statistics (mean, variance, frequency) & Activation statistics ($\mathbb{E}[a']$, $\text{Var}[a']$, frequency) on sample $X_S$ & Assumption: functionally similar neurons exhibit similar activation profiles. \newline \textbf{Drawback:} Fluctuations in statistics due to small weight changes or choice/size of $X_S$; activation “flickering”. \\
\hline
Optimal assignment-based methods (for alignment) & Weight similarity or activation correlation between neurons of two networks & Find an optimal permutation to match neurons between networks. \newline \textbf{Drawback (for single network canonization):} Require a second (reference) network; do not provide autonomous ordering. \\
\hline
Combinatorial structure analysis (sign patterns) & Characteristics of linear regions/sign patterns & Deep theoretical foundation. \newline \textbf{Drawback:} High computational complexity; sensitivity to changes in combinatorial structure. \\
\hline
Pruning/stable neuron identification & Weight magnitude, activation stability & Ranking by “importance” or stability. \newline \textbf{Drawback:} May remove functionally important neurons; activation stability depends on dataset. \\
\hline
\end{tabular}
\end{table}

\subsection{Key drawback of existing methods – instability}
\label{subsec:key_drawback_instability}

Despite their intuitive appeal and ease of implementation, many traditional neuron ordering methods reviewed in Subsection~\ref{subsec:traditional_ordering_alignment_methods} suffer from a serious fundamental drawback – \textbf{instability to small perturbations of network parameters}. This means that even minor changes in weights or biases, which can occur during a few training iterations, due to numerical computation inaccuracies, or when comparing two very similar but not identical networks, can lead to significant and unpredictable changes in the established neuron order.

Most of these methods rely on precise numerical values of parameters (weights, biases) or activation statistics, which are inherently extremely sensitive to the slightest changes inevitably arising during stochastic training, minor architectural modifications, or even numerical computation peculiarities \cite{mirzadeh2020understanding_regimes}. When sorting criterion values for different neurons are close, which is often observed in wide layers of modern networks, their relative order becomes highly unstable. The slightest parameter perturbation can radically change a neuron's position in the ordered list because many such heuristics lack a natural “margin” \cite{sartor2025advancing}. For example, if two neurons have almost identical bias values, their order can easily flip after one gradient descent step.

Activation statistic sensitivity is also significant. Neuron activation patterns, especially for inputs near their separating hyperplanes, can “flicker” (i.e., activation sign may fluctuate) with slight weight changes \cite{serra2021scaling, zhang2018efficient}. This, in turn, leads to instability in any derived statistics such as mean activation or its variance.

Such instability in the established neuron order has serious negative consequences:
\begin{itemize}
    \item It practically \textbf{invalidates attempts at component-wise comparison} of different networks, as “neuron $i$” in one model may have no meaningful correspondence to “neuron $i$” in another, even if the models are functionally very close \cite{ainsworth2022git}.
    \item \textbf{Training dynamics analysis} and tracking specialization of individual neurons over time become unreliable if their “identity” defined by ordering constantly changes.
    \item \textbf{Model interpretability} is reduced, as attributing specific functions to neurons based on their unstable positions or labels becomes challenging \cite{sharkey2025open}.
    \item Tasks requiring precise neuron matching, such as \textbf{model merging}, face insurmountable difficulties if the underlying ordering or alignment is unstable \cite{tatro2020optimizing, ainsworth2022git}.
\end{itemize}
Therefore, to develop truly useful tools for neural network analysis and comparison, ordering or characterization methods are needed that are significantly more robust to small parameter variations. This is one of the key motivations for the approach proposed in this work.

\section{Theoretical approach to neuron characterization: the concept of activation region ($\ARj$)}
\label{sec:theoretical_approach_AR}

After the parameters of hidden layer neurons have been brought to L2-normalized form (as described in Section~\ref{subsec:geometry_interpretation_l2}), eliminating scaling ambiguity, we can proceed to a deeper characterization of their individual functional contributions. This section introduces the fundamental concept of a neuron's “activation region” in the continuous input space. This concept will serve as a theoretical basis for developing neuron comparison methods and constructing a similarity metric, later approximated using sampled data.

\subsection{Definition of the “activation region”}
\label{subsec:definition_AR}

Consider neuron $j$ from an arbitrary hidden layer $k$. After L2 normalization, its parameters are represented by a normalized weight vector $W'_{k,j}$ (where $||W'_{k,j}||_2 = 1$, except for neurons with an originally zero weight vector) and a normalized bias $b'_{k,j}$. The pre-activation (logit) of this neuron for an input vector $x \in \Rspace{D_{in}^{(k)}}$ is $z'_{k,j}(x) = W'_{k,j} \cdot x + b'_{k,j}$.

The \textbf{activation region $\ARj$} of neuron $j$ is defined as the set of all points $x$ in the input space for which the neuron's pre-activation is strictly positive:
\[ \ARj = \{x \in \Rspace{D_{in}^{(k)}} \mid W'_{k,j} \cdot x + b'_{k,j} > 0\} \]
Geometrically, $\ARj$ is an \textbf{open half-space} in $\Rspace{D_{in}^{(k)}}$. Its boundary is the hyperplane $H_j$ defined by the equation $W'_{k,j} \cdot x + b'_{k,j} = 0$.

For practical analysis and subsequent matching with sampled data, it often makes sense to consider this region not in the entire infinite space $\Rspace{D_{in}^{(k)}}$ but within some \textbf{bounded region $K$} corresponding to the typical range or distribution of network inputs. Such a region $K$ can be, for example, a \textbf{hypercube} minimally enclosing all vectors from the training or test set or defined based on prior knowledge of input feature ranges.

Thus, the \textbf{effective activation region of neuron $j$ within $K$} is defined as the intersection, or restriction of $\ARj$ to $K$, denoted as $\ARrestricted{\ARj}{K}$:
\[ \ARrestricted{\ARj}{K} = \ARj \cap K = \{x \in K \mid W'_{k,j} \cdot x + b'_{k,j} > 0\} \]
This restriction to $K$ makes the activation region a bounded set, simplifying theoretical discussions of comparing such region volumes.

\begin{figure}[h!]
    \centering
    \includegraphics[width=0.7\textwidth]{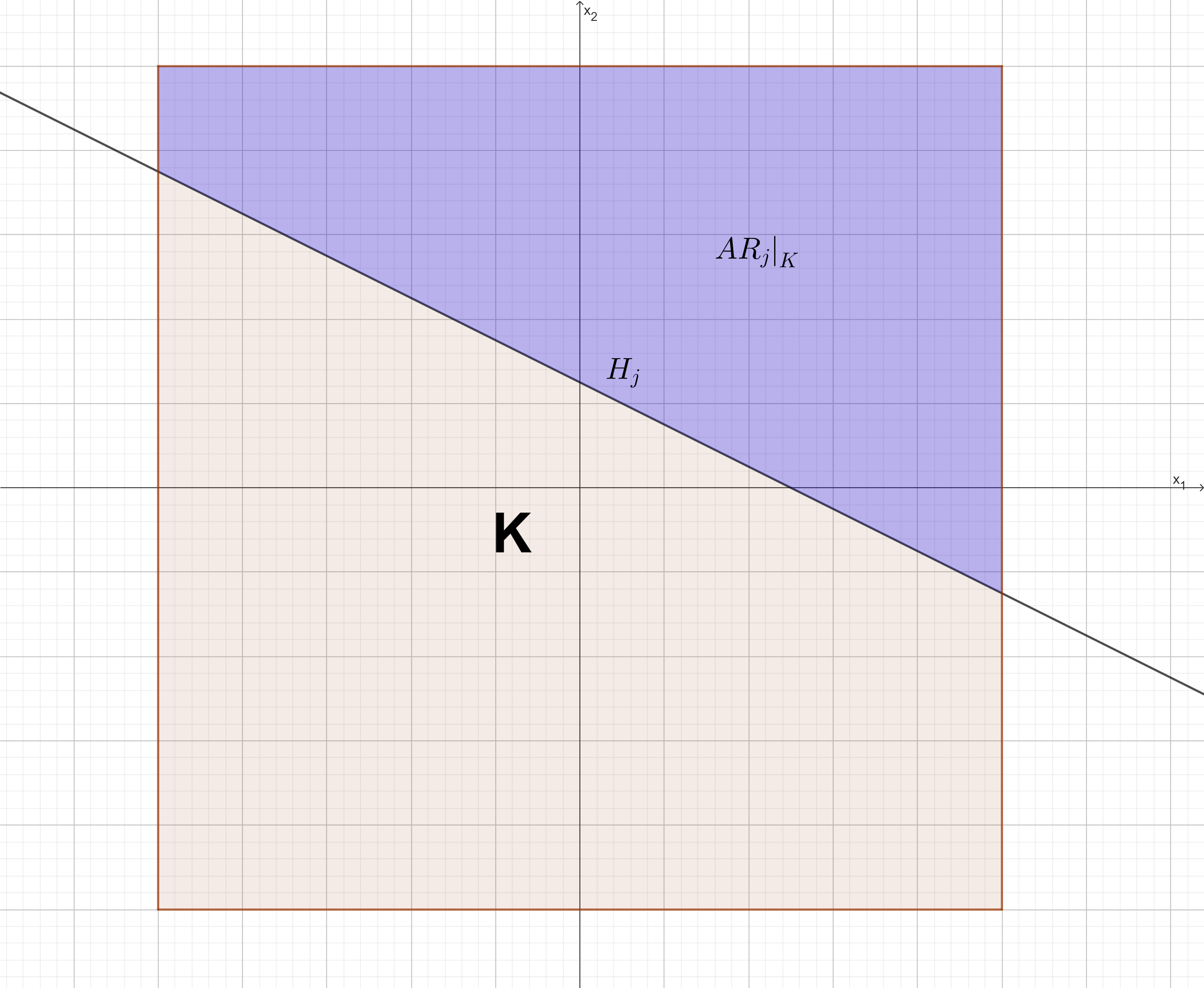}
    \textit{\caption{Illustration of neuron $j$’s activation region restricted to a bounded area $K$ ($\ARrestricted{\ARj}{K}$). The separating hyperplane $H_j$ and the shaded active region within $K$ are shown.}}
    \label{fig:activation_region_illustration}
\end{figure}

\subsection{Properties and geometric interpretation of \texorpdfstring{$\ARj$}{ARj (Activation Region j)}}
\label{subsec:properties_AR}

The activation region $\ARj$ defined in Subsection~\ref{subsec:definition_AR} (and in particular its restriction $\ARrestricted{\ARj}{K}$ to a bounded region $K \subset \Rspace{D_{in}^{(k)}}$) has several important properties and a clear geometric interpretation, key for understanding neuron roles and for subsequent similarity metric definitions.

\begin{itemize}
    \item \textbf{Geometric shape and convexity.}
    In unbounded space $\Rspace{D_{in}^{(k)}}$, the activation region $\ARj = \{x \in \Rspace{D_{in}^{(k)}} \mid W'_{k,j} \cdot x + b'_{k,j} > 0\}$ is an \textbf{open half-space}. Half-spaces are convex sets. When considered within a bounded convex region $K$ (e.g., a hypercube encompassing relevant input data), the \textbf{restricted activation region $\ARrestricted{\ARj}{K} = \ARj \cap K$} is also a \textbf{convex set}, since the intersection of convex sets is convex. Geometrically, $\ARrestricted{\ARj}{K}$ is a (possibly empty or degenerate) \textbf{convex polytope}. Its faces include portions of the separating hyperplane $H_j$ inside $K$ and parts of $K$’s own faces lying within $\ARj$.

    \item \textbf{Separating hyperplane $H_j$.}
    The boundary of $\ARj$ is the hyperplane $H_j$, defined by the linear part of the pre-activation:
    \[ H_j: W'_{k,j} \cdot x + b'_{k,j} = 0 \]
    As established in Subsection~\ref{subsec:geometry_interpretation_l2}:
    \begin{itemize}
        \item The weight vector $W'_{k,j}$ is a \textbf{unit normal vector} to hyperplane $H_j$, uniquely defining its orientation in $\Rspace{D_{in}^{(k)}}$. The direction of $W'_{k,j}$ points towards the positive half-space, i.e., towards $\ARj$ itself.
        \item The absolute value of the normalized bias $|b'_{k,j}|$ equals the \textbf{Euclidean distance from the origin to hyperplane $H_j$}. The sign of $b'_{k,j}$ determines the origin’s position relative to $H_j$ and $\ARj$:
            \begin{itemize}
                \item If $b'_{k,j} > 0$, then for $x=0$ we have $W'_{k,j} \cdot 0 + b'_{k,j} = b'_{k,j} > 0$, thus the origin lies in $\ARj$. Here, $H_j$ does not contain the origin and lies at distance $|b'_{k,j}|$ from it in the opposite direction of $W'_{k,j}$.
                \item If $b'_{k,j} < 0$, the origin does not belong to $\ARj$. $H_j$ lies at distance $|b'_{k,j}|$ from the origin in the direction of $W'_{k,j}$.
                \item If $b'_{k,j} = 0$, $H_j$ passes through the origin, which in this case lies on the boundary of $\ARj$ (but not in the open region $\ARj$ itself).
            \end{itemize}
    \end{itemize}

    \item \textbf{Effect of $(W'_{k,j}, b'_{k,j})$ on $\ARj$:}
    Changing L2-normalized neuron parameters has the following geometric effects:
    \begin{itemize}
        \item Changing the orientation of the unit normal vector $W'_{k,j}$ (with fixed $||W'_{k,j}||_2 = 1$ and $b'_{k,j}$) causes a \textbf{rotation} of hyperplane $H_j$ \textbf{around the origin}, with its distance $|b'_{k,j}|$ remaining unchanged.
        \item Changing $b'_{k,j}$ causes a \textbf{parallel shift} of $H_j$ (and the entire half-space $\ARj$) along its normal $W'_{k,j}$.
    \end{itemize}
    These changes directly affect the shape, size, and position of the restricted region $\ARrestricted{\ARj}{K}$.

    \item \textbf{Volume of the restricted activation region $\vol(\ARrestricted{\ARj}{K})$.}
    The restricted activation region $\ARrestricted{\ARj}{K}$, being a bounded polytope in $\Rspace{D_{in}^{(k)}}$ (assuming $K$ is bounded), has a finite $D_{in}^{(k)}$-dimensional volume denoted $\vol(\ARrestricted{\ARj}{K})$. This volume depends on $(W'_{k,j}, b'_{k,j})$, defining $\ARj$, and on the shape, size, and position of $K$. The volume can serve as an integral neuron characteristic, indicating what fraction of the “relevant” input space ($K$) it activates. Possible extremes include:
    \begin{itemize}
        \item $\vol(\ARrestricted{\ARj}{K}) = 0$: Occurs if $\ARj$ does not intersect the interior of $K$ (e.g., $K$ lies entirely in the “inactive” half-space $\{x \mid W'_{k,j} \cdot x + b'_{k,j} \leq 0\}$) or if the intersection has dimension less than $D_{in}^{(k)}$. Here, neuron $j$ is effectively never active for $x \in K$.
        \item $\vol(\ARrestricted{\ARj}{K}) = \vol(K)$: Occurs if $K$ is entirely contained within $\ARj$ ($K \subseteq \ARj$). Here, neuron $j$ is always active for all $x \in K$.
    \end{itemize}

    \item \textbf{Role in network function formation.}
    Each separating hyperplane $H_j$ contributes to the overall piecewise-linear partitioning of the input space $\Rspace{D_{in}^{(k)}}$ characterizing the ReLU network’s function. The arrangement of all such hyperplanes from all neurons (and layers) defines a complex combinatorial structure of convex polyhedral regions (often called linear regions, chambers, or “tops”) where the network function up to a layer is affine. The activation region $\ARj$ is one of the fundamental “building blocks” of this partitioning.
\end{itemize}
Understanding these properties and the geometric interpretation of $\ARj$ (and $\ARrestricted{\ARj}{K}$) is an important step for subsequent analysis of their stability to small parameter perturbations and for developing neuron comparison methods based on functional behavior in input space.

\subsection{Activation region $\ARj$ as a fundamental basis for neuron characterization}
\label{subsec:AR_as_signature_basis}

The concept of the activation region $\ARj$, defined via the L2-normalized neuron parameters $(W'_{k,j}, b'_{k,j})$, is not just an abstract geometric construct but a fundamental characteristic suitable for building a unique neuron “signature” or descriptor. Several key properties justify its suitability:

\begin{enumerate}
    \item \textbf{Functional determinacy and uniqueness:}
    The activation region $\ARj = \{x \in \Rspace{D_{in}^{(k)}} \mid W'_{k,j} \cdot x + b'_{k,j} > 0\}$ precisely defines the part of the input space to which the neuron responds positively (i.e. is “activated”). This partitioning of the space into active and inactive zones is the primary function of a ReLU neuron (before its interactions with other neurons and layers). Any two neurons with identical activation regions $\ARj$ are indistinguishable in their basic response to any input. Thus, $\ARj$ carries complete information about the threshold function implemented by the neuron.

    \item \textbf{Geometric interpretability:}
    As shown in Subsection~\ref{subsec:properties_AR}, $\ARj$ is an open half-space whose boundary hyperplane $H_j$ is uniquely determined by the unit normal vector $W'_{k,j}$ and the signed distance $|b'_{k,j}|$ from the origin. This clear geometric interpretation allows neurons to be analyzed and compared in terms of the orientation and position of their decision boundaries in feature space.

    \item \textbf{Theoretical robustness to small parameter perturbations:}
    The position and orientation of hyperplane $H_j$, and thus $\ARj$, depend continuously on the parameters $(W'_{k,j}, b'_{k,j})$. Small parameter changes cause small, predictable geometric changes (minor rotations and shifts of $H_j$). This means $\ARj$ as a geometric object does not undergo abrupt, catastrophic changes under small parameter variations. This continuous dependence favorably contrasts with direct comparisons of numerical weight or bias values, which can be more sensitive to minor fluctuations, especially when different neurons have similar values.

    \item \textbf{Informative for neuron comparison:}
    Comparing activation regions $\ARj$ and $\AR{l}$ of two neurons (or the same neuron at different training stages) reveals their functional similarity. The degree of overlap (e.g., the volume of $(\ARrestricted{\ARj}{K}) \cap (\ARrestricted{\AR{l}}{K})$ within a common bounded region $K$) or the closeness of their hyperplanes $H_j$ and $H_l$ can serve as measures of such similarity.

    \item \textbf{Basis for practically computable descriptors:}
    Although the continuous region $\ARj$ (or even $\ARrestricted{\ARj}{K}$) in high-dimensional space is difficult to use directly as a “signature” due to computational challenges in working with volumes and shapes in high dimensions, its fundamental properties (especially robustness and functional determinacy) make it an ideal \textbf{theoretical basis}. Building on $\ARj$, practically realizable discrete approximations or features (such as “sampled activation regions” or their hashed representations, discussed later) can inherit these positive properties.
\end{enumerate}

Thus, the activation region $\ARj$ is not just one possible way to describe a neuron but its essential characteristic, with the necessary geometric rigor and theoretical robustness. This makes $\ARj$ a starting point for developing reliable neuron identification, labeling, and comparison methods to overcome representation ambiguity in neural networks.

\subsection{Introducing a metric on the space of activation regions}
\label{subsec:metric_continuous_AR}

Building on the geometric concept of a neuron's activation region $\ARj$ and its restriction to a bounded convex region $K$, $\ARrestricted{\ARj}{K}$, we can define a \textbf{metric} to quantify differences between two neurons $j$ and $l$ of the same layer based on their activation regions. A natural approach is to use the \textbf{Jaccard distance} based on volumes of the respective regions.

Let $\ARrestricted{\ARj}{K}$ and $\ARrestricted{\AR{l}}{K}$ be the restricted activation regions of neurons $j$ and $l$, where $K \subset \Rspace{D_{in}^{(k)}}$ is a fixed bounded convex region with positive Lebesgue measure $\mu(K) > 0$. The Jaccard index $J(\ARrestricted{\ARj}{K}, \ARrestricted{\AR{l}}{K})$ is defined as:
\[ J(\ARrestricted{\ARj}{K}, \ARrestricted{\AR{l}}{K}) = \frac{\mu(\ARrestricted{\ARj}{K} \cap \ARrestricted{\AR{l}}{K})}{\mu(\ARrestricted{\ARj}{K} \cup \ARrestricted{\AR{l}}{K})} \]
assuming $\mu(\ARrestricted{\ARj}{K} \cup \ARrestricted{\AR{l}}{K}) > 0$. (If $\mu(\ARrestricted{\ARj}{K} \cup \ARrestricted{\AR{l}}{K}) = 0$, then $J=1$ if both volumes are zero, and $J=0$ otherwise, requiring detailed definitions to avoid division by zero).

The \textbf{Jaccard distance} $d_J$ between neurons $j$ and $l$ is then:
\[ d_J(j, l) \triangleq d_J(\ARrestricted{\ARj}{K}, \ARrestricted{\AR{l}}{K}) = 1 - J(\ARrestricted{\ARj}{K}, \ARrestricted{\AR{l}}{K}) \]
This function $d_J$ maps a neuron pair (via their activation regions) to $[0,1]$.

\paragraph{Properties of $d_J$ as a metric.}
The function $d_J$ satisfies metric axioms on the set of all Lebesgue-measurable subsets of $K$ (where elements are the restricted activation regions $\ARrestricted{\ARj}{K}$):
\begin{enumerate}
    \item \textbf{Non-negativity:} $d_J(j, l) \ge 0$.
    \item \textbf{Identity of indiscernibles:} $d_J(j, l) = 0 \iff \mu(\ARrestricted{\ARj}{K} \Delta \ARrestricted{\AR{l}}{K}) = 0$, where $\Delta$ is symmetric difference (i.e., regions are identical up to measure zero).
    \item \textbf{Symmetry:} $d_J(j, l) = d_J(l, j)$.
    \item \textbf{Triangle inequality:} For any neurons $j, l, m$, $d_J(j, m) \le d_J(j, l) + d_J(l, m)$.
\end{enumerate}
Thus, $(\mathcal{A}_K, d_J)$, where $\mathcal{A}_K$ is the set of all possible restricted activation regions $\ARrestricted{\ARj}{K}$ within $K$, forms a \textbf{metric space}. This enables quantitative assessment of differences between neurons based on activation region geometry.

\paragraph{Computational complexity.}
Despite its correctness as a metric, \textbf{direct computation of $d_J$ in high-dimensional spaces is very computationally expensive}. This is due to the need to determine geometries and compute volumes (Lebesgue measures) of high-dimensional polytopes $\ARrestricted{\ARj}{K}$, $\ARrestricted{\AR{l}}{K}$, as well as their intersections and unions. In high dimensions ($D_{in}^{(k)}$), exact volume computation is \#P-hard for general polytopes.

Due to these computational barriers, the continuous $d_J$ metric cannot be directly used in practical algorithms. Nevertheless, it serves as an important \textbf{theoretical foundation and benchmark}. Understanding its properties and limitations motivates developing discrete approximations based on sampled data, which will be covered in Section~\ref{sec:sampled_AR_practical_estimation}.

\section{Practical estimation of activation regions: neuron $j$’s activation region restricted to sample $\XS$ ($\ARrestricted{\ARj}{\XS}$)}
\label{sec:sampled_AR_practical_estimation}

In the previous section (Section~\ref{sec:theoretical_approach_AR}) we introduced the concept of a neuron's activation region $\ARj$ as a fundamental geometric characteristic and discussed its theoretical properties, including robustness and its suitability as a similarity metric basis. However, as noted, working directly with continuous regions $\ARj$ (even restricted to a bounded $K$) in high dimensions is computationally prohibitive, making analytical or exact numerical volume or intersection calculations infeasible for real-sized neural networks.

To overcome these limitations and transition to practically feasible neuron analysis and comparison methods, we introduce the concept of the \textbf{activation region restricted to a sample} and its equivalent vector representation — the \textbf{sampled activation signature}. This approach approximates neuron behavior on a finite, representative set of input data rather than on the entire continuous space.

\subsection{Definition of activation region restricted to sample ($\ARrestricted{\ARj}{\XS}$) and its vector representation ($S_j$)}
\label{subsec:definition_SAR_Sj}

Let $\XS = \{\xs{1}, \xs{2}, \dots, \xs{\Nsamp}\}$ be a fixed \textbf{sample} of $\Nsamp$ input vectors, each $\xs{s} \in \Rspace{D_{in}^{(k)}}$ (typically $\xs{s} \in K$, where $K$ is the bounded relevant input region defined in Subsection~\ref{subsec:definition_AR}).

For neuron $j$ of layer $k$ with L2-normalized parameters $(W'_{k,j}, b'_{k,j})$, its \textbf{activation region restricted to sample $\XS$}, denoted $\ARrestricted{\ARj}{\XS}$, is the subset of $\XS$ where neuron $j$ is active:
\[ \ARrestricted{\ARj}{\XS} = \{\xs{s} \in \XS \mid W'_{k,j} \cdot \xs{s} + b'_{k,j} > 0\} \]
Thus, $\ARrestricted{\ARj}{\XS}$ is the discrete set of sample points “falling inside” the neuron's continuous activation region $\ARj$.

For computational purposes and for subsequent hashing and comparison, this discrete characteristic is conveniently represented as a \textbf{binary vector}, termed the \textbf{sampled activation signature (SAS)} of neuron $j$, denoted $\Sj$. This vector has length $\Nsamp$:
\[ \Sj = [s_{j,1}, s_{j,2}, \dots, s_{j,\Nsamp}] \]
where each element is defined by:
\[ s_{j,s} = \indicate{W'_{k,j} \cdot \xs{s} + b'_{k,j} > 0} = \begin{cases} 1 & \text{if } W'_{k,j} \cdot \xs{s} + b'_{k,j} > 0 \\ 0 & \text{otherwise} \end{cases} \]
Here, $\indicate{\cdot}$ is the indicator function. Thus, $\Sj$ is effectively the neuron’s “fingerprint” or activation profile over the sample $\XS$, indicating for which inputs it is active.

\subsection{Transition from geometric model to data: approximating $\ARj$ properties}
\label{subsec:geometric_to_data_transition}

Introducing the sampled activation signature $\Sj$ (representing $\ARrestricted{\ARj}{\XS}$) is a key step enabling the transition from theoretical analysis of continuous activation regions $\ARj$ to practical estimation based on finite data samples. This not only resolves computational complexity issues in high dimensions but also yields neuron characteristics directly reflecting behavior on task-relevant data.

\paragraph{Approximating $\ARj$ properties via $\Sj$.}
\begin{itemize}
    \item \textbf{Representing the “active zone”:} $\Sj$ shows which of the $\Nsamp$ sample points fall within $\ARj$. Although it does not describe the entire infinite region, it captures its “trace” or “projection” onto the sample $\XS$. If $\XS$ is representative of the data distribution, $\Sj$ effectively reflects $\ARj$’s behavior in the most important input areas.

    \item \textbf{Estimating “massiveness” or “influence” of $\ARj$:} Instead of computing the intractable $\vol(\ARrestricted{\ARj}{K})$, a simpler proxy is the fraction of sample points activating neuron $j$: $P_j = \frac{1}{\Nsamp} \sum_{s=1}^{\Nsamp} s_{j,s}$, indicating how “large” or “influential” $\ARj$ is relative to the sample.

    \item \textbf{Comparing neurons via their signatures:} Comparing neurons $j$ and $l$ reduces to comparing their sampled activation signatures $\Sj$ and $S_l$, which is far easier than comparing continuous half-spaces.
\end{itemize}

\paragraph{Transforming the similarity metric.}
Most importantly, theoretical metrics such as the Jaccard distance $d_J(\ARrestricted{\ARj}{K}, \ARrestricted{\AR{l}}{K})$ (discussed in Subsection~\ref{subsec:metric_continuous_AR}) based on volumes can now be replaced with discrete analogues computed directly on $\Sj$ and $S_l$.

Viewing $\Sj$ and $S_l$ as binary indicators of sample membership in activation regions, the Jaccard index for these vectors (or the sets of indices of active samples) is:
\[ J(\Sj, S_l) = \frac{|\{s \mid s_{j,s} = 1 \text{ and } s_{l,s} = 1\}|}{|\{s \mid s_{j,s} = 1 \text{ or } s_{l,s} = 1\}|} = \frac{\sum_{s=1}^{\Nsamp} s_{j,s} \cdot s_{l,s}}{\sum_{s=1}^{\Nsamp} \max(s_{j,s}, s_{l,s})} \]
where the numerator counts samples activating both neurons (analogue of intersection volume), and the denominator counts samples activating at least one neuron (analogue of union volume).

Thus, the \textbf{discrete Jaccard distance} $d_J^S(j,l)$ between neurons $j$ and $l$ based on their sampled activation signatures is:
\[ d_J^S(j,l) = 1 - J(\Sj, S_l) \]

This metric $d_J^S(j,l)$:
\begin{enumerate}
    \item \textbf{Is computable:} Requiring only bitwise operations on vectors of length $\Nsamp$.
    \item \textbf{Is interpretable:} Still reflecting the degree of non-overlap of the two neurons’ “active zones” over the data.
    \item \textbf{Is a metric} on the space of binary vectors (or their equivalent sets).
\end{enumerate}

Thus, sampling and the transition to sampled activation signatures $\Sj$ transform theoretically grounded but computationally infeasible concepts of continuous $\ARj$ analysis into practically applicable and computable analogues. The approximation quality and reliability of conclusions drawn from $\Sj$ depend directly on the sample $\XS$’s representativeness and size, discussed in subsequent subsections.

\subsection{Sample size estimation ($\Nsamp$)}
\label{subsec:sample_size_estimation}

Choosing an adequate sample size $\Nsamp$ (number of input vectors $\xs{s}$ in $\XS$) is critical for obtaining reliable and informative sampled activation signatures $\Sj$. Too small a sample may not reflect true neuron behavior and lead to unstable results, while an excessively large sample increases computational costs. The goal is to find $\Nsamp$ providing sufficient “resolution” to distinguish functionally different neurons with stable characteristic estimates.

Here we consider theoretical approaches to estimating $\Nsamp$, adapted for a network with $D_{in}^{(k)} = 512$ inputs and $N_k = 2048$ neurons in the analyzed layer.

\subsubsection{Theoretical foundations: Vapnik–Chervonenkis theory}
\label{subsubsec:vc_theory_foundations}

To justify sample size choice, we refer to VC-theory \cite{vapnik1971uniform, vapnik1998statistical}, which provides rigorous sample complexity bounds in statistical learning.

\paragraph{Key VC-theory concepts.}
Developed by Vladimir Vapnik and Alexey Chervonenkis (1960–1990), it studies conditions under which empirical frequencies uniformly converge to true probabilities \cite{vapnik1971uniform, vapnik1998statistical}. The central concept is the \textbf{VC-dimension} of a function class $\mathcal{H}$: the largest $n$ such that $\mathcal{H}$ can “shatter” any set of $n$ points, realizing all $2^n$ dichotomies.

The fundamental connection between PAC learnability and VC-dimension was established by Blumer et al. \cite{blumer1989learnability}: a concept class is PAC-learnable iff its VC-dimension is finite.

For the class of oriented hyperplanes in $\mathbb{R}^d$, VC-dimension is $d+1$ \cite{vapnik1998statistical, sontag1998vc_dimension_neural_networks}. This follows from showing a set of size $d+1$ can be shattered, while Radon’s theorem implies no set of size $d+2$ can be \cite{devroye1996probabilistic, mohri2018foundations, shalev2014understanding, kearns1994introduction}.

\paragraph{Uniform convergence theorem.}
The classic result \cite{vapnik1971uniform} states: for a class $\mathcal{C}$ with finite VC-dimension $d_{VC}$ and i.i.d. samples $x^{(1)}, \ldots, x^{(N)}$ from $P$,
\begin{equation}
\label{eq:vc_uniform_convergence}
\lim_{N \to \infty} \sup_{C \in \mathcal{C}} \left| \frac{1}{N} \sum_{i=1}^N \mathbf{1}_{C}(x^{(i)}) - P(C) \right| = 0 
\end{equation}
where $\mathbf{1}_{C}(\cdot)$ is the indicator function. Thus, empirical frequencies converge uniformly to true probabilities as $N \to \infty$.

\subsubsection{Application to neuron characterization}
\label{subsubsec:vc_application_neurons}

\paragraph{Problem formalization.}
In our context, define $\mathcal{H} = \{h_1, h_2, \ldots, h_{N_k}\}$, where each $h_j: \Rspace{D_{in}^{(k)}} \to \{0,1\}$ is:
\begin{equation}
h_j(x) = \indicate{W'_{k,j} \cdot x + b'_{k,j} > 0}.
\end{equation}
Each $h_j$ indicates neuron $j$’s activation region $\ARj$. We want empirical activation frequency
\begin{equation}
\hat{P}_j = \frac{1}{\Nsamp} \sum_{s=1}^{\Nsamp} s_{j,s} = \frac{1}{\Nsamp} \sum_{s=1}^{\Nsamp} h_j(\xs{s})
\end{equation}
to estimate true activation probability $P_j = P_X(\ARj)$ within tolerance $\varepsilon$ and confidence $1-\delta$ for all $N_k$ neurons.

\paragraph{Accounting for multiple hypotheses.}
We analyze $N_k$ fixed neurons. For one neuron, VC-dimension is $d = D_{in}^{(k)} + 1 = 513$.

\subsubsection{Sample complexity bounds}
\label{subsubsec:sample_complexity_bounds}

\paragraph{Single hypothesis bound.}
For one binary function $h$ with VC-dimension $d$, to ensure with probability $1-\delta'$ that $|P_{N}(h) - P(h)| \leq \varepsilon$, sample size $N$ must satisfy:
\begin{equation}
N \geq \frac{C_1}{\varepsilon^2}\left(d \log\left(\frac{C_2 N}{d}\right) + \log\left(\frac{C_3}{\delta'}\right)\right),
\end{equation}
where $C_1 \approx 1, C_2 \approx 2e, C_3 \approx 2$.

\paragraph{Multiple hypotheses bound.}
We require $|\hat{P}_j - P_j| \leq \varepsilon$ simultaneously for all $j=1,\ldots,N_k$ with confidence $1-\delta$. Applying union bound replaces $\delta'$ with $\delta/N_k$, yielding:
\begin{equation}
\label{eq:vc_bound_multiple_hypotheses_final}
\Nsamp \geq \frac{1}{\varepsilon^2}\left((D_{in}^{(k)}+1) \log\left(\frac{2e\Nsamp}{D_{in}^{(k)}+1}\right) + \log\left(\frac{2N_k}{\delta}\right)\right).
\end{equation}
This transcendental equation in $\Nsamp$ is solved numerically.

\paragraph{Numerical calculation for our system.}
With $D_{in}^{(k)}=512$, $N_k=2048$, $\varepsilon=0.05$ (5
\begin{equation}
\Nsamp \geq 400\left(513 \log\left(\frac{2e\Nsamp}{513}\right) + \log(409600)\right).
\end{equation}
Iterative solutions yield:
\begin{equation}
\boxed{\Nsamp \geq 2,054,825.}
\end{equation}

\subsubsection{Interpretation of VC estimate and practical choice of $\Nsamp$}
\label{subsubsec:vc_interpretation_practical_choice}

The VC-theory-based estimate $\Nsamp \geq 2,054,825$ (for $D_{in}^{(k)} = 512$, $N_k = 2048$, $\varepsilon = 0.05$, $\delta = 0.01$) implies: with $\ge 99\%$ confidence, \textbf{each} of the 2048 neurons’ empirical activation frequency $\hat{P}_j$ differs from its true activation probability $P_j$ by at most 5\%. This ensures high statistical reliability of sampled activation signatures $\Sj$ as characteristics of the true activation regions $\ARj$ under data distribution $P_X$.

Key points:

\begin{enumerate}
    \item \textbf{Focus on individual measure accuracy:}
    Guarantees precise estimation of each $\ARj$’s “size” relative to data distribution.
    \item \textbf{Does not guarantee combinatorial structure coverage:}
    It ensures individual region frequency accuracy, not sampling all linear regions (chambers) formed by joint hyperplane intersections.
    \item \textbf{Potential conservatism:}
    VC bounds are mathematically tight but often conservative in practice, especially with union-bound correction for $N_k$ hypotheses.
\end{enumerate}

Despite this, VC-based sample sizing remains a rigorous approach for statistical validity of $\Sj$. Given the theoretical $\approx 2 \times 10^6$ estimate and computational constraints, choosing $\Nsamp$ in the $2$ million range balances accuracy and resource use. Exact choice depends on desired statistical guarantees vs. available compute. The sampling strategy for $\XS$ is equally important and discussed next.

\subsection{Sampling strategies for $\XS$}
\label{subsec:sampling_strategies}

The properties of $\XS$ (distribution, generation method) directly impact $\Sj$ representativeness and analysis stability. Sample formation involves a trade-off among computational efficiency, input space coverage, real data relevance, and theoretical guarantees.

Here we discuss three main groups of approaches.

\subsubsection{Using existing data}
\label{sssec:existing_data_sampling}

\paragraph{Method.}
Extract $\Nsamp$ examples from existing datasets (train, validation, test) by random or stratified sampling.

\paragraph{Advantages.}
Yields $\Sj$ reflecting neuron behavior on practically relevant inputs. Since samples match real data distribution, results have direct task-specific interpretation \cite{montavon2017methods}. Also computationally efficient, requiring no synthetic data generation.

\paragraph{Disadvantages.}
Potential dataset bias may fail to cover important input regions. Results become dataset-specific and may not capture “absolute” neuron properties. Redundant similar samples can reduce analysis informativeness.

\subsubsection{Random sampling from input space}
\label{sssec:random_sampling}

\paragraph{Method.}
Generate points $\xs{s}$ randomly from a chosen distribution (uniform within a hypercube covering input ranges, or multivariate normal fitted to data).

\paragraph{Advantages.}
Provides broader, less biased coverage of input space. Useful for probing “absolute” neuron properties independent of dataset specifics, potentially revealing unexpected activation patterns in underrepresented regions.

\paragraph{Disadvantages.}
Many generated points may lie in low-probability or irrelevant regions for real data, leading to inefficient sampling of decision boundaries and $\Sj$ that poorly reflect practical neuron behavior.

\subsubsection{Specialized structured sampling methods}
\label{sssec:structured_sampling}

Beyond simple random sampling, advanced methods aim for more efficient, uniform coverage of high-dimensional input space. These are useful for obtaining representative activation signatures $\Sj$ with limited sample size $\Nsamp$.

\paragraph{Latin Hypercube Sampling (LHS).}
\subparagraph{Method:}
LHS divides each input dimension’s range into $\Nsamp$ equally probable strata, then samples exactly one point from each stratum along each dimension \cite{mckay1979comparison}, ensuring uniform univariate projections.

\subparagraph{Advantages:}
Provides better axis-aligned coverage than pure random sampling at the same $\Nsamp$ \cite{mckay1979comparison}, effective for studying effects of individual features or low-order interactions.

\subparagraph{Disadvantages:}
More complex implementation; ensures good marginal coverage but not optimal high-order multidimensional interactions; may induce structural patterns.

\subparagraph{Practical implementation:}
Available in Python via \texttt{scipy.stats.qmc.LatinHypercube} (e.g., \texttt{sampler = qmc.LatinHypercube(d=D\_in)} and \texttt{sampler.random(n=Nsamp)}), also in SALib.

\paragraph{Low-Discrepancy Sequences.}
\subparagraph{Method:}
Quasi-random sequences (e.g. Sobol \cite{sobol1967distribution, antonov1979implementation}, Halton, Faure \cite{niederreiter1988low}) minimize discrepancy, filling space more evenly \cite{owen2003quasi}.

\subparagraph{Advantages:}
Better convergence rates ($O(1/\Nsamp)$ vs $O(1/\sqrt{\Nsamp})$ for Monte Carlo); more uniform coverage in high dimensions \cite{owen2003quasi}. Sobol sequences perform well up to thousands of dimensions.

\subparagraph{Disadvantages:}
Require understanding of parameters (e.g. scrambling, skipping initial points); above $d \gg 100$ advantages diminish; Halton may degrade due to high-order correlations. Some sequences perform best when $\Nsamp$ is a power of two.

\subparagraph{Practical implementation:}
Available in Python via \texttt{scipy.stats.qmc.Sobol} (e.g., \texttt{sampler = qmc.Sobol(d=D\_in, scramble=True)}) and \texttt{Halton}, with sampling via \texttt{sampler.random(n=Nsamp)}.

\subsubsection{Comparative analysis and recommendations}
\label{sssec:sampling_recommendations}

For architectural analysis and understanding general neuron behavior, structured methods (LHS or low-discrepancy) are recommended for systematic coverage.

For application-focused analysis (practical network behavior on real data), sampling from existing datasets is preferred \cite{montavon2017methods}.

For comprehensive analysis, combine both: real data for relevance, structured sampling for broad coverage. Choice depends on research goals and compute resources.

\section{Analysis of Neuron Characteristics Based on the Sample and Their Comparison}
\label{sec:analysis_sampled_characteristics}

After defining the concept of the activation region of a neuron restricted to a sample ($\ARrestricted{\ARj}{\XS}$) and its vector representation as a sampled activation signature ($\Sj$) in Section~\ref{sec:sampled_AR_practical_estimation}, and discussing the formation of the sample $\XS$, we now turn to methods for analysing these discrete characteristics. The central object at this stage is the matrix aggregating the signatures of all neurons in the analysed layer.

\subsection{Formation of the Sampled Activation Signature Matrix (<<Neurons × Samples>>)}
\label{subsec:activation_matrix_formation}

The main result of the previous stage (Section~\ref{sec:sampled_AR_practical_estimation}) is the set of sampled activation signatures (SAS) $\Sj$ for each neuron $j$ in the considered layer $k$. Each signature $\Sj$ is a binary vector of length $\Nsamp$, whose elements $s_{j,s}$ indicate whether ($s_{j,s}=1$) or not ($s_{j,s}=0$) neuron $j$ is activated for the $s$-th input sample $\xs{s}$ from the sample $\XS$.

For further analysis, these individual signatures $\Sj$ are conveniently organised into a single matrix, which we will call the \textbf{sampled activation signature matrix} (or, for brevity, the <<Neurons × Samples>> matrix), denoted by $M_{act}$.

\begin{itemize}
    \item \textbf{Structure of the matrix $M_{act}$:}
    \begin{itemize}
        \item Each \textbf{row} of $M_{act}$ corresponds to a neuron in layer $k$. If the layer has $N_k$ neurons, the matrix will have $N_k$ rows. The $j$-th row of the matrix is the sampled activation signature vector $\Sj$ of neuron $j$.
        \item Each \textbf{column} of $M_{act}$ corresponds to one input sample from $\XS$. If the sample contains $\Nsamp$ examples, the matrix will have $\Nsamp$ columns.
        \item \textbf{Matrix element $(M_{act})_{j,s}$} equals $s_{j,s}$, i.e. a binary value (0 or 1) indicating the activation state of neuron $j$ for sample $s$.
    \end{itemize}
    \item \textbf{Matrix dimensions:} $M_{act}$ has size $N_k \times \Nsamp$. Given our system parameters ($N_k = 2048$ neurons, $\Nsamp \approx 2 \cdot 10^6$ samples, according to the estimate in Subsection~\ref{subsubsec:vc_interpretation_practical_choice}), this matrix is quite large.
    \item \textbf{Formation process:}
    To form $M_{act}$, the following steps are performed:
    \begin{enumerate}
        \item The L2-normalised parameters $(W'_{k,j}, b'_{k,j})$ are used for each neuron $j$ in layer $k$.
        \item For each neuron $j \in \{1, \dots, N_k\}$ and each input sample $\xs{s} \in \XS$ (where $s \in \{1, \dots, \Nsamp\}$), the pre-activation is computed:
        \[ z'_{k,j}(\xs{s}) = W'_{k,j} \cdot \xs{s} + b'_{k,j} \]
        \item The matrix element $(M_{act})_{j,s}$ is defined as:
        \[ (M_{act})_{j,s} = s_{j,s} = \indicate{z'_{k,j}(\xs{s}) > 0} \]
    \end{enumerate}
\end{itemize}
The resulting binary matrix $M_{act}$ is a key informational resource for all subsequent analysis stages, including the identification of specific types of neurons (e.g. always active or inactive neurons for this sample), the discovery of functionally similar (or identical on the sample) neurons, and, most importantly, for quantitatively evaluating pairwise similarity between neurons. This matrix will also form the basis for constructing compact hashed representations of neurons, discussed in Section~\ref{sec:hashing_methods}.

\subsection{Primary analysis of the sampled activation signature matrix}
\label{subsec:primary_analysis_activation_matrix}

\paragraph{Introduction.}
The sampled activation signature matrix $M_{act}$ with size $N_k \times \Nsamp$ contains exhaustive information about the response of each analysed neuron to each sample in $\XS$. Primary analysis of this matrix allows not only basic behaviour characteristics to be extracted but also preliminary classification and filtering of neurons. This greatly simplifies subsequent detailed comparative analysis aimed at forming unique and informative signatures for neurons actually involved in processing information on the sample.

This section considers three main types of neurons with <<extreme>> behaviour on the sample $\XS$: inactive (<<dead>>), always active, and functionally identical neurons. Identifying and appropriately handling such neurons is an important preliminary step before building complex characteristics and similarity metrics.

\subsubsection{Identification of inactive (<<dead>>) neurons}
\label{sssec:dead_neurons}

\paragraph{Definition and detection criterion.}
A neuron $j$ is considered inactive (or <<dead>>) on the sample $\XS$ if all elements of its signature $\Sj$ are zero:
\[ \sum_{s=1}^{\Nsamp} s_{j,s} = 0 \]
This means that for all $\xs{s} \in \XS$, the condition $W'_{k,j} \cdot \xs{s} + b'_{k,j} \leq 0$ holds.

\paragraph{Interpretation and impact on signature formation.}
The presence of inactive neurons may indicate the <<dying ReLU problem>> \cite{goodfellow2016deep, nair2010rectified, xu2015empirical}. Causes may include poor initialisation, overly high learning rates, or data characteristics not covering the neuron's activation region.
From a signature formation perspective, \textbf{a dead neuron has a trivial signature $\Sj = \vec{0}$} on the sample, carrying no information about its separating properties. Such a signature will be identical for all dead neurons, making them indistinguishable based on it.

\paragraph{Practical actions.}
\begin{itemize}
    \item \textbf{Documentation:} Record the number and proportion of inactive neurons as a characteristic of the network or sample.
    \item \textbf{Exclusion from unique signature analysis:} As their signatures are trivial, such neurons are \textbf{excluded from further processes of building unique characteristics} and pairwise comparison. They are labelled as a separate category <<dead on $X_S$>>.
    \item \textbf{Diagnostic value:} A high proportion of dead neurons (e.g. >10-40\%) may indicate training issues or unrepresentative $\XS$ \cite{karpathy2015cs231n}.
\end{itemize}

\subsubsection{Identification of always active neurons}
\label{sssec:always_active_neurons}

\paragraph{Definition and detection criterion.}
A neuron $j$ is considered always active on the sample $\XS$ if all elements of its signature $\Sj$ are equal to one:
\[ \sum_{s=1}^{\Nsamp} s_{j,s} = \Nsamp \]
This means that for all $\xs{s} \in \XS$, the condition $W'_{k,j} \cdot \xs{s} + b'_{k,j} > 0$ holds.

\paragraph{Interpretation and impact on signature formation.}
An always active neuron indicates that the entire sample $\XS$ lies within its activation region $\ARj$. For ReLU activation, such a neuron acts as a linear element on this sample \cite{goodfellow2016deep, he2015delving}. Its signature $\Sj = \vec{1}$ is also \textbf{trivial in terms of discriminative power on $\XS$} and will be identical for all always active neurons.

\paragraph{Practical actions.}
\begin{itemize}
    \item \textbf{Documentation:} Record their count.
    \item \textbf{Exclusion from unique signature analysis:} Similar to dead neurons, always active neurons on $\XS$ are \textbf{excluded from the main process of constructing unique signatures}, as their <<signature>> carries no information for discrimination.
    \item \textbf{Diagnostic information:} A high proportion of such neurons may indicate insufficient variability in the sample $\XS$ or specific properties of the trained network.
\end{itemize}

\subsubsection{Identification of identical (on sample \texorpdfstring{$\XS$}{XS}) neurons}
\label{sssec:identical_neurons_on_sample}

\paragraph{Definition and detection criterion.}
Two or more neurons (e.g., $j$ and $l$), that are neither dead nor always active, are considered identical on the sample $\XS$ if their sampled activation signatures are identical:
\[ \Sj = S_l \quad \Leftrightarrow \quad s_{j,s} = s_{l,s} \text{ for all } s = 1, 2, \ldots, \Nsamp \]

\paragraph{Interpretation and impact on uniqueness of signatures.}
Identical non-trivial signatures $\Sj = S_l$ indicate indistinguishable behaviour on all samples from $\XS$. This may suggest functional redundancy on the given sample \cite{schmidhuber2000neural} or strong correlation. For constructing unique signatures, \textbf{neurons with identical $\Sj$ will initially have the same <<raw>> signature}.

\paragraph{Clustering algorithm and practical actions.}
To ensure uniqueness of subsequent characteristics:
\begin{itemize}
    \item \textbf{Clustering:} Neurons with identical non-trivial signatures $\Sj$ are grouped into equivalence classes.
    \item \textbf{Representative selection:} For each class, one representative neuron is chosen (e.g. with the smallest original index). \textbf{Only these representatives will have further unique signatures constructed and analysed}. The other neurons in the group are marked as equivalent on $\XS$.
    \item \textbf{Documentation:} Record the sizes of these groups.
\end{itemize}

\subsubsection{Summary and preparation for detailed analysis}
\label{sssec:summary_primary_analysis_for_signatures}

The results of the primary analysis of the matrix $M_{act}$ form the basis for filtering neurons prior to constructing their unique signatures.
\begin{itemize}
    \item Neurons identified as \textbf{inactive} or \textbf{always active} on the sample $\XS$ are \textbf{excluded from further analysis}, as their signatures ($\vec{0}$ and $\vec{1}$, respectively) are trivial and carry no unique discriminative information on $\XS$.
    \item For groups of \textbf{identical (on $\XS$) neurons}, only one representative is selected. Further unique signatures are constructed only for these representatives, and the other neurons in such groups are considered to be represented by this signature.
\end{itemize}
Such preliminary processing allows focusing efforts on neurons demonstrating non-trivial and diverse behaviour on the sample, which is necessary for obtaining meaningful and distinct unique characteristics while reducing computational complexity in subsequent stages.

\subsection{Quantitative assessment of neuron similarity based on sampled activation signatures \texorpdfstring{$\Sj$}{Sj}}
\label{subsec:quantitative_similarity_Sj}

After constructing the matrix of sampled activation signatures $M_{act}$ (Section~\ref{subsec:activation_matrix_formation}) and performing its preliminary analysis to identify and process neurons with trivial or identical signatures on the sample $\XS$ (Section~\ref{subsec:primary_analysis_activation_matrix}), we proceed to quantitatively assess pairwise similarity between neurons demonstrating non-trivial behaviour. This assessment is based on their sampled activation signature vectors $\Sj$, which represent a discrete approximation of the activation regions $\ARj$ on the specific sample $\XS$.

As previously noted in Section~\ref{subsec:geometric_to_data_transition}, for comparing binary vectors $\Sj$ and $S_l$ (where $j$ and $l$ are indices of two neurons in the considered layer, and $\Sj, S_l \in \{0,1\}^{\Nsamp}$), the \textbf{Jaccard index} is a natural measure. For two sampled activation signatures $\Sj$ and $S_l$, it is defined as the ratio of the number of shared active samples to the number of samples where at least one neuron is active:
$$J(\Sj, S_l) = \frac{|\{s \mid s_{j,s} = 1 \text{ and } s_{l,s} = 1\}|}{|\{s \mid s_{j,s} = 1 \text{ or } s_{l,s} = 1\}|}$$
In terms of vector operations, if $s_{j,s}$ and $s_{l,s}$ are the components of vectors $\Sj$ and $S_l$, this formula can be written as:
$$J(\Sj, S_l) = \frac{\sum_{s=1}^{\Nsamp} s_{j,s} \cdot s_{l,s}}{\sum_{s=1}^{\Nsamp} \max(s_{j,s}, s_{l,s})}$$
provided that at least one of the neurons is active on at least one sample (i.e. the denominator is not zero). If both vectors $\Sj$ and $S_l$ are zero vectors (which corresponds to two <<dead>> neurons on the sample $\XS$, already filtered at Section~\ref{sssec:dead_neurons}), then $J(\Sj, S_l)$ can be defined as 1 (since they are identical). If one vector is zero and the other is not, then $J(\Sj, S_l)=0$.

The Jaccard index $J(\Sj, S_l)$ takes values in the range $[0,1]$:
\begin{itemize}
    \item $J(\Sj, S_l) = 1$ indicates that neurons $j$ and $l$ have identical sampled activation signatures (i.e. $\Sj = S_l$), activating on exactly the same subset of samples from $\XS$.
    \item $J(\Sj, S_l) = 0$ indicates that the sets of samples on which neurons $j$ and $l$ activate do not overlap (i.e. there is no sample $\xs{s} \in \XS$ on which both are active).
\end{itemize}

Based on the Jaccard index, the \textbf{discrete Jaccard distance} $d_J^S(j,l)$ between neurons $j$ and $l$ is defined as:
$$d_J^S(j,l) = 1 - J(\Sj, S_l)$$
This distance also takes values in the range $[0,1]$ and has the properties of a metric: $d_J^S(j,l) = 0$ if and only if $\Sj = S_l$, $d_J^S(j,l) = d_J^S(l,j)$, and it satisfies the triangle inequality.

Computing pairwise Jaccard distances $d_J^S(j,l)$ for all (or for an interesting subset of) neurons in the layer allows:
\begin{enumerate}
    \item Constructing a pairwise distance (or similarity) matrix, providing a complete view of the functional proximity between neurons based on their responses on $\XS$.
    \item Identifying clusters of functionally similar neurons.
    \item This distance matrix will serve as the basis for constructing the cost matrix $\Cuv$ in the neuron matching algorithm between two different layers (Section~\ref{sec:optimal_neuron_matching}).
\end{enumerate}

Despite its computational simplicity and interpretability for comparing two vectors $\Sj$, direct calculation of all pairwise similarities for a large number of neurons $N_k$ (requiring $O(N_k^2 \cdot \Nsamp)$ operations) can be expensive if $\Nsamp$ is also large. This motivates the use of hashing methods such as Bloom filters and MinHash (discussed in Sections~\ref{sec:hashing_methods}) for efficient approximation of the Jaccard index and acceleration of the comparison process.

\section{Efficient Representation of Sampled Activation Regions: Hashing Methods}
\label{sec:hashing_methods}

After constructing the sampled activation signatures $\Sj$ for the neurons of the analysed layer (Section~\ref{sec:analysis_sampled_characteristics}), we obtain informative but large representations of each neuron's behaviour on the sample $\XS$. Each signature $\Sj$ is a binary vector of length $\Nsamp$, which, at theoretically justified sample sizes (on the order of $2 \cdot 10^6$ elements, see Section~\ref{subsubsec:vc_interpretation_practical_choice}), creates significant computational and resource constraints for practical application of the proposed approach.

The key challenge is that constructing similarity metrics between layers requires computing pairwise distances between neurons, involving comparison of their sampled activation signatures. However, direct methods for comparing such long binary vectors are computationally infeasible for real-world tasks. This motivates the search for efficient compressed representations of signatures.

The solution lies in the area of \textbf{Locality-Sensitive Hashing (LSH)} — a family of algorithmic techniques specifically designed to produce compressed representations (hashes) that form a new metric space. The key property of LSH is that the similarity metric in this hash space inherits the properties of the original metric (in our case, the Jaccard index for vectors $\Sj$) with predictable accuracy. This allows distances or similarities to be computed between compact hashes instead of the large original objects. In this section, we review the theoretical foundations of such methods and their adaptation to the neural network analysis task.

\subsection{The scalability problem when comparing long binary vectors \texorpdfstring{$\Sj$}{Sj}}
\label{subsec:scalability_problem_Sj_comparison}

The sampled activation signatures $\Sj$ obtained in previous sections provide an accurate discrete approximation of neuron activation regions $\ARj$ on the sample $\XS$. The larger the sample size $\Nsamp$, the higher the approximation accuracy and the statistical reliability of inferences about neuron behaviour. However, increasing $\Nsamp$ leads to a substantial rise in computational costs:

\begin{itemize}
    \item \textbf{Computational complexity of pairwise comparison:} For $N_k$ neurons in a layer, $\binom{N_k}{2}$ pairwise distances must be computed. Each comparison of two vectors $\Sj$ and $S_l$ requires $O(\Nsamp)$ operations to compute the Jaccard index (Section~\ref{subsec:quantitative_similarity_Sj}). The total complexity is $O(N_k^2 \cdot \Nsamp)$. With realistic parameters ($N_k = 2048$, $\Nsamp \approx 2 \cdot 10^6$), this amounts to approximately $8 \cdot 10^{12}$ elementary operations.
    \item \textbf{Memory requirements:} Storing all activation signatures for one layer ($N_k=2048, \Nsamp = 2 \cdot 10^6$) requires $N_k \times \Nsamp$ bits, amounting to approximately $512$ MB.
\end{itemize}
These constraints make the direct approach impractical.

\subsection{Principles of LSH: Metric inheritance and LSH-preserving transformations}
\label{sssec:lsh_principles_metric_inheritance}

The search for efficient representations of vectors $\Sj$ is not merely a matter of dimensionality reduction. It is crucial that these compressed representations (hashes or sketches) preserve or approximate information about the original similarity metric (in our case, the Jaccard index). Operations in the hash space must meaningfully reflect the proximity relations present in the original signature space.

This task is addressed by \textbf{Locality Sensitive Hashing (LSH)} methods \cite{indyk1998approximate, leskovec2014mining}. Unlike cryptographic hash functions, which deliberately minimise correlations between input and output (avalanche effect), LSH functions are designed so that similar objects in the original space are mapped to identical or <<nearby>> hash values with high probability.

The key idea of LSH is to construct a mapping $H: \mathcal{S} \to \mathcal{S}'$ from the original signature space $(\mathcal{S}, d)$ to the sketch space $(\mathcal{S}', \rho)$ such that the metric $\rho$ in the sketch space \textbf{inherits} the properties of the original metric $d$ in a predictable manner. This means that if two signatures $S_a, S_b$ are close in $(\mathcal{S}, d)$, then their images $H(S_a), H(S_b)$ will also be close in $(\mathcal{S}', \rho)$. More formally, for ranking and comparison tasks, an essential property is \textbf{(probabilistic) preservation of relative distance orderings}: if for four signatures $S_a, S_b, S_c, S_d \in \mathcal{S}$ we have $d(S_a, S_b) < d(S_c, S_d)$, then it should hold that:
$$\Pr[\rho(H(S_a), H(S_b)) < \rho(H(S_c), H(S_d))] \geq p$$
where $p$ is the probability ($p > 0.5$) characterising the reliability of order preservation.

The ability of LSH functions to provide such metric inheritance and order preservation is closely related to the concept of \textbf{LSH-preserving transformations}, formally characterised, for example, in Chierichetti and Kumar (2015) \cite{chierichetti2015lsh}, which guarantee that fundamental LSH properties (the ability to distinguish near and far objects) are not lost under mapping. Mappings with the order-preservation property are also known in the broader context as \textbf{order-preserving embeddings} or \textbf{isotonic mappings}.

The practical value of this approach is that it allows distances or similarity measures computed on compact and computationally efficient sketches to be used as reliable proxies for the corresponding measures in the original high-dimensional space. This is the key to overcoming the scalability problem in neuron signature analysis.

\subsection{MinHash function}
\label{ssec:minhash_function}

\subsubsection{Definition}
\label{sssec:minhash_definition}

MinHash is a dimensionality reduction method for sets, enabling efficient estimation of their similarity. Let there be a universal set $U$. For any subset $S \subseteq U$ and a random permutation $\pi$ of the elements of $U$, the MinHash function $h_{\pi}(S)$ is defined as the minimum value taken by $\pi$ on the elements of $S$:
$$h_{\pi}(S) = \min_{x \in S} \pi(x)$$
Thus, $h_{\pi}(S)$ is a "signature" or "sketch" of the set $S$.

In the context of binary vectors, a vector $V$ of length $N$ is represented as a set $S_V = \{i \mid V[i]=1, 1 \leq i \leq N\}$, that is, the set of indices where the vector has a value of one. The universe $U$ in this case is the set of all indices $\{1, \dots, N\}$, and $\pi$ is a random permutation of these indices. Then, MinHash for vector $V$ is computed as $h_{\pi}(V) = \min_{i \in S_V} \pi(i)$.

\subsubsection{Computation algorithm (conceptual)}
\label{sssec:minhash_computation_conceptual}

To compute $h_{\pi}(S)$ for a given set $S$:
\begin{enumerate}
    \item Apply the random permutation $\pi$ to all elements of the universe $U$. Each element $x \in U$ is assigned its rank (value) $\pi(x)$.
    \item Consider only those elements that belong to set $S$.
    \item The value of $h_{\pi}(S)$ is the smallest rank among the elements belonging to $S$.
\end{enumerate}
In practice, for large universes $U$ (e.g., for long binary vectors), explicitly applying permutations is computationally expensive. Instead, families of universal hash functions $\mathcal{H}: U \to \mathbb{N}$ are used (where $\mathbb{N}$ is the set of natural numbers, or a sufficiently large range). The MinHash value is approximated as $h_{H}(S) = \min_{x \in S} H(x)$. Using different hash functions $H_1, H_2, \dots, H_K$ from such a family is equivalent to using different random permutations.

\paragraph{Example of MinHash computation for a binary vector}
\label{par:minhash_example_binary_vector}

Consider a specific example of MinHash computation for a binary vector.
Let the vector length be $N=6$. The universe of indices is $U=\{1,2,3,4,5,6\}$.
Suppose we have the following random permutation $\pi$ of these indices:
\begin{align*}
    \pi(1) &= 3 \\
    \pi(2) &= 1 \\
    \pi(3) &= 6 \\
    \pi(4) &= 2 \\
    \pi(5) &= 4 \\
    \pi(6) &= 5
\end{align*}
Let the binary vector be $V=[1,0,0,1,1,0]$.

Computation process:
\begin{enumerate}
    \item \textbf{Determine the set of indices with ones:} \\
    For vector $V$, the set of indices where $V[i]=1$ is $S_V = \{1,4,5\}$.
    \item \textbf{Find the permutation values for these indices:}
    \begin{itemize}
        \item For index $1 \in S_V$, permutation value is $\pi(1)=3$.
        \item For index $4 \in S_V$, permutation value is $\pi(4)=2$.
        \item For index $5 \in S_V$, permutation value is $\pi(5)=4$.
    \end{itemize}
    \item \textbf{Compute MinHash:} \\
    The MinHash value for vector $V$ with permutation $\pi$ is the minimum of the obtained permutation values:
    $$h_{\pi}(V) = \min\{\pi(1),\pi(4),\pi(5)\} = \min\{3,2,4\} = 2$$
\end{enumerate}
Thus, for vector $V=[1,0,0,1,1,0]$ and the given permutation $\pi$, the MinHash value $h_{\pi}(V)=2$.

\subsubsection{Property for Jaccard coefficient}
\label{sssec:minhash_jaccard_property}

The key property of the MinHash function is its direct relation to the Jaccard coefficient. For two sets $A$ and $B$ (or their corresponding binary vectors) and a randomly chosen permutation $\pi$, the probability of their MinHash values matching is equal to the Jaccard coefficient $J(A,B)$:
$$\Pr[h_{\pi}(A)=h_{\pi}(B)] = J(A,B) = \frac{|A \cap B|}{|A \cup B|}$$
This fundamental property (proved by Broder in 1997 \cite{broder1997resemblance}) underlies the use of MinHash for similarity estimation.

\subsection{Jaccard distance estimation based on MinHash}
\label{ssec:jaccard_distance_estimation_minhash}

To improve the accuracy of estimating the Jaccard coefficient (or distance), not just one, but a family of $K$ independent MinHash functions is used: $h_1, h_2, \dots, h_K$. Each function $h_t$ is generated based on its own independent random permutation $\pi_t$ (or independent universal hash function $H_t$).

The Jaccard distance $d_J(A,B)$ is defined as $d_J(A,B)=1-J(A,B)$. Its unbiased estimator $\rho(A,B)$ can be obtained based on $K$ MinHash functions as follows:
$$\rho(A,B) = \frac{1}{K} \sum_{t=1}^{K} \mathbb{1}[h_t(A) \neq h_t(B)]$$
where $\mathbb{1}[\cdot]$ is the indicator function (equals 1 if the condition is true, and 0 otherwise).

The expectation of this estimator equals the true Jaccard distance:
\begin{align*}
\mathbb{E}[\rho(A,B)] &= \mathbb{E}\left[\frac{1}{K} \sum_{t=1}^{K} \mathbf{1}[h_t(A) \neq h_t(B)]\right] \\
&= \frac{1}{K} \sum_{t=1}^{K} \mathbb{E}[\mathbf{1}[h_t(A) \neq h_t(B)]]
\end{align*}
Since $\mathbb{E}[\mathbb{1}[h_t(A) \neq h_t(B)]] = \Pr[h_t(A) \neq h_t(B)] = 1 - J(A,B) = d_J(A,B)$, then
$$\mathbb{E}[\rho(A,B)] = \frac{1}{K} \sum_{t=1}^{K} d_J(A,B) = d_J(A,B)$$
Thus, $\rho(A,B)$ is an unbiased estimator of $d_J(A,B)$.

\subsection{Preservation of Jaccard distance ordering}
\label{ssec:jaccard_order_preservation}

\subsubsection{Theorem statement}
\label{sssec:order_preservation_theorem_statement}

Let there be four sets (or binary vectors) $S_i, S_j, S_k, S_l$ \textbf{from universe $U$ of size $N$}. Denote their true Jaccard distances as $d_J(S_i, S_j)$ and $d_J(S_k, S_l)$. Assume the inequality $d_J(S_i, S_j) < d_J(S_k, S_l)$ holds. Define $\delta = d_J(S_k, S_l) - d_J(S_i, S_j) > 0$.

If the distance estimates use the metric $\rho(A,B) = \frac{1}{K}\sum_{t=1}^{K} \mathbb{1}[h_t(A) \neq h_t(B)]$, based on $K$ independent MinHash functions, then the probability of preserving the ordering of these distances is lower-bounded as follows:
$$\Pr[\rho(S_i, S_j) < \rho(S_k, S_l)] \geq 1 - 2\exp\left(-\frac{K\delta^2}{2}\right)$$
\textbf{provided that:}
\begin{enumerate}
    \item The hash functions $h_1, \ldots, h_K$ are independent random permutations of the universe $U$
    \item $\delta \geq \frac{1}{N}$ (minimum distinguishable distance for the given universe)
\end{enumerate}

\subsubsection{Proof}
\label{sssec:order_preservation_theorem_proof}

\paragraph{Notation:}
Let $X = \rho(S_i, S_j)$. Then $\mu_X = \mathbb{E}[X] = d_J(S_i, S_j)$.
Let $Y = \rho(S_k, S_l)$. Then $\mu_Y = \mathbb{E}[Y] = d_J(S_k, S_l)$.
By the theorem condition, $\mu_X < \mu_Y$, and $\delta = \mu_Y - \mu_X > 0$.

\paragraph{Condition for order preservation:} The inequality $X<Y$ is guaranteed to hold if the following two conditions are met simultaneously:
\begin{align}
X &< \mu_X + \delta/2 \label{eq:condX} \\
Y &> \mu_Y - \delta/2 \label{eq:condY}
\end{align}
Since $\mu_X + \delta/2 = \mu_X + (\mu_Y - \mu_X)/2 = (\mu_X + \mu_Y)/2$, and $\mu_Y - \delta/2 = \mu_Y - (\mu_Y - \mu_X)/2 = (\mu_X + \mu_Y)/2$, both conditions \eqref{eq:condX} and \eqref{eq:condY} compare $X$ and $Y$ to the same midpoint $(\mu_X + \mu_Y)/2$. If $X$ is to the left of this point and $Y$ is to the right, then $X<Y$.

\paragraph{“Bad” events:} Consider the events that can violate these favorable conditions:
\begin{itemize}
    \item $E_1: X \geq \mu_X + \delta/2$ (equivalent to $X - \mu_X \geq \delta/2$).
    \item $E_2: Y \leq \mu_Y - \delta/2$ (equivalent to $Y - \mu_Y \leq -\delta/2$, or $\mu_Y - Y \geq \delta/2$).
\end{itemize}

\paragraph{Applying Hoeffding’s inequality:}
The estimator $\rho(A,B)$ is the average of $K$ independent Bernoulli random variables $\mathbb{1}[h_t(A) \neq h_t(B)]$, each taking values in $\{0,1\}$. For such averages, the one-sided Hoeffding inequality for any $\varepsilon > 0$ states:
$$\Pr[\rho(A,B) - \mathbb{E}[\rho(A,B)] \geq \varepsilon] \leq \exp(-2K\varepsilon^2)$$
$$\Pr[\rho(A,B) - \mathbb{E}[\rho(A,B)] \leq -\varepsilon] \leq \exp(-2K\varepsilon^2)$$
Applying this to events $E_1$ and $E_2$ with $\varepsilon = \delta/2$:
\begin{align*}
\Pr[E_1] &= \Pr[X - \mu_X \geq \delta/2] \leq \exp(-2K(\delta/2)^2) = \exp\left(-\frac{K\delta^2}{2}\right) \\
\Pr[E_2] &= \Pr[Y - \mu_Y \leq -\delta/2] \leq \exp(-2K(\delta/2)^2) = \exp\left(-\frac{K\delta^2}{2}\right)
\end{align*}

\paragraph{Using the union bound:}
The probability that at least one of the “bad” events $E_1$ or $E_2$ occurs is bounded by the sum of their probabilities:
$$\Pr[E_1 \cup E_2] \leq \Pr[E_1] + \Pr[E_2] \leq \exp\left(-\frac{K\delta^2}{2}\right) + \exp\left(-\frac{K\delta^2}{2}\right) = 2\exp\left(-\frac{K\delta^2}{2}\right)$$

\paragraph{Conclusion:}
If neither $E_1$ nor $E_2$ occurs, then conditions $X < \mu_X + \delta/2$ and $Y > \mu_Y - \delta/2$ hold, which in turn guarantees $X<Y$. The probability that neither $E_1$ nor $E_2$ occurs is $1 - \Pr[E_1 \cup E_2]$.
Therefore,
$$\Pr[X<Y] \geq 1 - \Pr[E_1 \cup E_2] \geq 1 - 2\exp\left(-\frac{K\delta^2}{2}\right)$$
Thus, $\Pr[\rho(S_i, S_j) < \rho(S_k, S_l)] \geq 1 - 2\exp\left(-\frac{K\delta^2}{2}\right)$.
The theorem is proven.

\subsubsection{Practical limitations and remarks}
\label{sssec:practical_limitations_remarks}

\textbf{Remark 1 (On independence of hash functions).} The requirement for independence of $K$ permutations imposes an implicit constraint $K \leq N!$. In practice, for large $N$ this constraint is negligible, but for small $N$ (on the order of hundreds) with $K > N$ some hash functions may become dependent.

\textbf{Remark 2 (On discreteness of the space).} For binary vectors of length $N$, the Jaccard coefficient can take no more than $N+1$ distinct values of the form $\frac{k}{m}$, where $k \leq m \leq N$. Consequently, the minimum nonzero value of $\delta$ is approximately $\frac{1}{N}$. This implies that:
\begin{itemize}
    \item For small $N$ (e.g., $N < 1000$) even small differences in Jaccard coefficients represent significant jumps in the discrete space
    \item The theorem is most informative when $\delta \gg \frac{1}{N}$
\end{itemize}

\textbf{Remark 3 (On choosing $K$ depending on $N$).} Although formally the probability bound does not depend on $N$, the practical choice of $K$ should consider:
\begin{itemize}
    \item \textbf{Discreteness of the estimate}: $\rho(A,B)$ takes values in increments of $\frac{1}{K}$
    \item \textbf{Resolution}: to distinguish distances differing by $\delta$, it is desirable that $\frac{1}{K} < \frac{\delta}{2}$
    \item \textbf{Computational efficiency}: for vectors of length $N \sim 10^6$, using $K \sim 10^2-10^3$ usually provides a good balance between accuracy and speed
\end{itemize}

\textbf{Remark 4 (On asymptotic accuracy).} The bound in the theorem becomes more accurate as $N \to \infty$ with fixed $K$, when discrete effects become negligible and the space of possible Jaccard coefficient values approaches the continuum $[0,1]$.

\subsection{Estimating parameters \texorpdfstring{$K$}{K} and \texorpdfstring{$\delta$}{delta} for given reliability}
\label{sssec:minhash_parameter_estimation_reliability}

The order preservation theorem (Subsection~\ref{sssec:order_preservation_theorem_statement}) gives us the following lower bound for the probability $P_s$ of correct order preservation of distances:
$$P_s = \Pr[\rho(S_i, S_j) < \rho(S_k, S_l)] \geq 1 - 2\exp\left(-\frac{K\delta^2}{2}\right)$$
where $K$ is the number of independent MinHash functions, and $\delta = d_J(S_k, S_l) - d_J(S_i, S_j) > 0$ is the minimum difference between two true Jaccard distances that we want to reliably distinguish.

Our goal is to ensure a given reliability (confidence) $P_s = 1 - \alpha$, where $\alpha$ is the acceptable error probability (e.g., $\alpha = 0.05$ for $95\%$ confidence).
From this, we require:
$$1 - 2\exp\left(-\frac{K\delta^2}{2}\right) \geq 1 - \alpha$$
This transforms to:
$$2\exp\left(-\frac{K\delta^2}{2}\right) \leq \alpha$$
$$\exp\left(-\frac{K\delta^2}{2}\right) \leq \frac{\alpha}{2}$$
Taking the natural logarithm of both sides (and changing the inequality sign since the logarithm is an increasing function and we multiply by -1):
$$-\frac{K\delta^2}{2} \leq \ln\left(\frac{\alpha}{2}\right)$$
$$\frac{K\delta^2}{2} \geq -\ln\left(\frac{\alpha}{2}\right) = \ln\left(\frac{2}{\alpha}\right)$$
Hence, the required number of MinHash functions $K$ is:
$$K \geq \frac{2\ln(2/\alpha)}{\delta^2}$$
Consider an example with required confidence $P_s = 0.95$, i.e., $\alpha = 0.05$.
Then $2/\alpha = 2/0.05 = 40$, and $\ln(40) \approx 3.689$.
Therefore, for $95\%$ confidence:
$$K \geq \frac{2 \times 3.689}{\delta^2} \approx \frac{7.378}{\delta^2}$$
This formula shows an important dependence:
\begin{itemize}
    \item The smaller the $\delta$ (i.e., the closer the Jaccard distances we need to distinguish), the larger the number of MinHash functions $K$ required to preserve their order with given confidence. For example:
    \begin{itemize}
        \item If we want to distinguish distance pairs differing by $\delta = 0.1$, then $K \geq \frac{7.378}{(0.1)^2} = \frac{7.378}{0.01} \approx 738$.
        \item If $\delta = 0.05$, then $K \geq \frac{7.378}{(0.05)^2} = \frac{7.378}{0.0025} \approx 2951$.
    \end{itemize}
    \item If the number of hash functions $K$ is limited (e.g., due to performance or sketch size considerations), then there exists a minimum value $\delta_{min}$ that can be reliably distinguished with confidence $1-\alpha$:
    $$\delta_{min} \geq \sqrt{\frac{2\ln(2/\alpha)}{K}}$$
\end{itemize}
Thus, choosing $K$ is a trade-off between the desired precision in distinguishing close distances (small $\delta$), required confidence ($1-\alpha$), and computational resources. In practice, $\delta$ is determined based on how fine the differences in neuron similarity need to be captured for analysis purposes.

\subsection{MinHash implementations in Python libraries}
\label{sssec:minhash_python_libraries}

MinHash is a well-known and widely used algorithm, and its implementations are available in several Python libraries, simplifying its practical application.

One of the most popular and specialized libraries for MinHash and LSH is \texttt{datasketch}\footnote{Detailed information and documentation on MinHash in the \texttt{datasketch} library are available at: \url{http://ekzhu.com/datasketch/minhash.html}.}. It provides:
\begin{itemize}
    \item Efficient MinHash implementations for creating set signatures (sketches).
    \item Functions for estimating the Jaccard index based on these signatures.
    \item LSH data structures, such as `MinHashLSH` (for similarity search based on the Jaccard index) and `MinHashLSHForest` (for more accurate search via additional indexing).
    \item Support for Weighted MinHash and other variations.
\end{itemize}
The `datasketch` library makes it easy to create MinHash objects, add set elements (or hashed values of elements for large data), and then compare them to obtain Jaccard index estimates. The number of hash functions ($K$, often the parameter \texttt{num\_perm} in `datasketch`) is configurable.

Conceptual usage example (not executable code, illustrates the idea):
\begin{verbatim}
from datasketch import MinHash

# Create MinHash objects with K=128 hash functions
m1 = MinHash(num_perm=128)
m2 = MinHash(num_perm=128)

# Add elements (e.g., indices of active samples)
for element in set_A:
    m1.update(element.encode('utf8')) # Elements must be byte strings
for element in set_B:
    m2.update(element.encode('utf8'))

# Estimate Jaccard index
jaccard_estimate = m1.jaccard(m2)

# Estimate Jaccard distance
distance_estimate = 1 - jaccard_estimate
\end{verbatim}

Besides `datasketch`, other implementations or tutorials are available in the public domain. However, `datasketch` is a good starting point for production use due to its optimization and feature set.

When using ready-made libraries, it is important to understand how exactly they implement the choice of hash functions (often these are not explicit permutations, but families of universal hash functions approximating the behavior of random permutations) and how parameters are configured to achieve the desired accuracy and performance.

\subsubsection{Summary of MinHash and its role as neuron signatures}
\label{sssec:minhash_summary_signatures}

As a result of the analysis conducted, we can conclude that the MinHash method provides an efficient mechanism for generating compact \textbf{signatures (sketches)} from the original, potentially very long, binary vectors of sampled neuron activations $\Sj$. These MinHash signatures have key properties that make them the desired tool for characterizing and distinguishing unique neurons within this work:

\begin{enumerate}
    \item \textbf{Ensuring uniqueness and distinguishability:} With a sufficient number of hash functions $K$, MinHash signatures will, with high probability, be distinct for neurons that have different (not completely overlapping) activation patterns on sample $\XS$. This allows them to be used for identifying and tracking functionally distinct neural units.

    \item \textbf{Basis for similarity/distance metrics:} As shown (Subsections~\ref{ssec:jaccard_distance_estimation_minhash} and \ref{ssec:jaccard_order_preservation}), MinHash signatures provide an unbiased estimate of the Jaccard index (and distance) between the original sets of active samples. Moreover, the relative ordering of these distances is preserved with provable probability. This means we can quantitatively measure and compare the functional closeness of neurons using their compact MinHash sketches.
\end{enumerate}

An important practical aspect is the \textbf{flexibility in choosing the MinHash signature length} (parameter $K$), which allows the method to be adapted to different tasks and accuracy requirements:

\begin{itemize}
    \item For tasks where a less precise but faster similarity estimate or simply tracking significant changes in neuron behavior is needed (e.g., analyzing training dynamics or identifying roughly similar neurons), \textbf{shorter MinHash signatures} (smaller $K$ values) can be used. This reduces computational costs and storage volume.
    \item For tasks requiring high comparison accuracy, such as detailed matching of two trained neural networks of the same architecture or precise measurement of functional similarity for building canonical forms, \textbf{longer MinHash signatures} (larger $K$ values) should be used. This provides a more accurate approximation of the Jaccard distance and more reliable order preservation, at the cost of somewhat higher computational resources.
\end{itemize}

Thus, MinHash provides us with a powerful and flexible tool for transitioning from bulky raw data on neuron activations to their compact yet informative representations, which will form the basis for further analysis steps, including constructing the neuron pairwise similarity matrix and ultimately defining the similarity metric between neural networks.

\section{Finding the optimal pairwise neuron matching}
\label{sec:optimal_neuron_matching}

After obtaining compact and informative representations (signatures) for each neuron based on the analysis of their sampled activation regions and hashing methods (Section~\ref{sec:hashing_methods}), the next step is to develop an algorithm to establish correspondence between neurons from two different layers. These layers may belong to different neural networks (with identical or similar architecture), or to the same network at different stages of training or after various transformations.

The goal of this section is to describe the procedure for finding such a pairwise matching that is optimal in terms of functional similarity of the matched neurons. This is a key stage for subsequent computation of similarity metrics between layers or entire networks, as well as for other analysis tasks such as knowledge transfer or model merging. The process consists of two main stages: first, constructing a matrix quantitatively describing the “cost” or “distance” of matching each possible pair of neurons from the two layers, and then applying an algorithm to find the optimal assignment based on this matrix.

\subsection{Constructing the cost (distance) matrix \texorpdfstring{$\Cuv$}{Cuv}}
\label{ssec:cost_matrix_construction}

The first step in the neuron matching algorithm is to quantitatively evaluate how “costly” or “undesirable” it is to match a specific neuron $u$ from the first analyzed layer (denote it Layer 1) with neuron $v$ from the second analyzed layer (Layer 2). This evaluation is represented as a cost matrix (or distance matrix) $C$.

Let Layer 1 contain $M$ neurons and Layer 2 contain $N$ neurons. Then the cost matrix $C$ has dimension $M \times N$. Each element of the matrix $C_{uv}$ (where $1 \le u \le M, 1 \le v \le N$) represents the computed cost or distance between neuron $u$ from Layer 1 and neuron $v$ from Layer 2. A low value of $C_{uv}$ corresponds to a high degree of similarity between neurons $u$ and $v$, while a high value indicates significant difference.

In the context of our work, neuron characteristics are based on their sampled activation signatures $\Sj$, which were transformed into compact MinHash sketches (signatures) in Section~\ref{sec:hashing_methods} (specifically, see discussion of MinHash in Subsection~\ref{ssec:minhash_function} and onwards). The measure of similarity between the original signatures $\Sj$ is the Jaccard index, and the distance measure is the Jaccard distance $d_J = 1 - J$.

As shown in Subsection~\ref{ssec:jaccard_distance_estimation_minhash}, the Jaccard distance $d_J(S_u, S_v)$ between two original signatures $S_u$ and $S_v$ can be efficiently estimated using their MinHash sketches $H(S_u)$ and $H(S_v)$ as:
$$ \rho(H(S_u), H(S_v)) = \frac{1}{K} \sum_{t=1}^{K} \mathbf{1}_{\{h_t(H(S_u)) \neq h_t(H(S_v))\}} $$
where $K$ is the number of MinHash functions in the sketch. This estimator $\rho(H(S_u), H(S_v))$ is an unbiased estimate of $d_J(S_u, S_v)$ and takes values in the range $[0, 1]$ (0 for identical sketches, 1 for completely different sketches in MinHash terms).

Thus, the matrix element $C_{uv}$ is defined as the estimated Jaccard distance between neuron $u$ (from Layer 1) and neuron $v$ (from Layer 2) based on their MinHash sketches:
$$ C_{uv} = \rho(H(S_u), H(S_v)) $$
where $H(S_u)$ and $H(S_v)$ are the MinHash sketches obtained for the sampled activation signatures $S_u$ and $S_v$ of the respective neurons.

The resulting matrix $C$ of size $M \times N$ contains non-negative elements, where each $C_{uv}$ quantitatively characterizes the degree of functional difference (estimated via their behavior on sample $\XS$) between neuron $u$ from the first layer and neuron $v$ from the second. This matrix will serve as input for the optimal assignment algorithm discussed in the next subsection.

\subsection{Applying the Hungarian algorithm (or similar) to find the optimal matching}
\label{ssec:hungarian_algorithm_application}

After constructing the cost matrix $C$ (where each element $\Cuv$ represents the distance or cost of matching neuron $u$ from Layer 1 with neuron $v$ from Layer 2) (Subsection~\ref{ssec:cost_matrix_construction}), the next task is to find the optimal one-to-one correspondence between neurons of these two layers. Optimality here means that the total cost (or total distance) of all selected neuron pairs is minimized. This problem is known as the \textbf{assignment problem} or the minimum weight bipartite matching problem.

\subsubsection{The assignment problem and the Hungarian algorithm}
\label{sssec:assignment_problem_hungarian}

Let Layer 1 contain $M$ neurons and Layer 2 contain $N$ neurons.
\begin{itemize}
    \item If $M=N$ (layers have the same number of neurons), the task is to find a permutation $p$ of indices $\{1, \dots, M\}$ that minimizes the total cost $\sum_{u=1}^{M} C_{u, p(u)}$. Each neuron from Layer 1 is matched to exactly one unique neuron from Layer 2.
    \item If $M \neq N$, say $M < N$, the task is to select $M$ different neurons from Layer 2 and match them to the $M$ neurons of Layer 1 to minimize the total cost. Each neuron from the smaller layer (Layer 1) will be matched, and from the larger layer (Layer 2) $M$ neurons are selected for matching. Similarly if $M > N$.
\end{itemize}

The classical and widely known method for solving the assignment problem for square cost matrices is the \textbf{Hungarian algorithm}, developed by Harold Kuhn in 1955 based on earlier work by Hungarian mathematicians Dénes Kőnig and Jenő Egerváry. The algorithm finds the optimal assignment in polynomial time, typically with computational complexity of $O(n^3)$ for an $n \times n$ matrix.

\paragraph{Principle of the Hungarian algorithm (conceptually):}
The Hungarian algorithm works by transforming the cost matrix and finding the optimal set of independent zeros. The main steps (high-level overview) are:
\begin{enumerate}
    \item \textbf{Matrix reduction:} Subtract the minimum element of each row from all elements of that row. Then subtract the minimum element of each column from all elements of that column. The goal is to create as many zeros as possible without changing the optimal solution.
    \item \textbf{Finding an optimal assignment among zeros:} Attempt to find $n$ independent zeros (i.e. $n$ zeros such that no two are in the same row or column). If successful, the optimal assignment is found.
    \item \textbf{Modifying the matrix (if optimal assignment not found):} If $n$ independent zeros cannot be found, modify the matrix by covering all zeros with the minimum number of horizontal and vertical lines. Then find the smallest element not covered by any line. Subtract this element from all uncovered cells and add it to all cells covered twice (i.e. intersections of the covering lines). This creates new zeros or changes relative costs while preserving optimality. Return to step 2.
\end{enumerate}
The process repeats until an optimal assignment is found.

Although the Hungarian algorithm was originally designed for square matrices, it can be adapted for rectangular $M \times N$ matrices. This is often done by padding the matrix to square with dummy neurons assigned very high matching costs, or modern implementations in libraries directly support rectangular matrices, finding the optimal assignment for $\min(M,N)$ pairs.

\subsubsection{Implementation and availability in libraries}
\label{sssec:hungarian_implementation_libraries}

Implementing the Hungarian algorithm from scratch is a non-trivial task. Fortunately, efficient and well-tested implementations are available in standard scientific libraries.

\paragraph{Python:}
In the Python ecosystem, the Hungarian algorithm (often referred to as the Munkres algorithm, after one of its efficient implementations) is available in the SciPy library:
\begin{itemize}
    \item The function \texttt{scipy.optimize.linear\_sum\_assignment(cost\_matrix)} takes a cost matrix as input and returns two arrays: \texttt{row\_ind} and \texttt{col\_ind}. Each pair \texttt{(row\_ind[i], col\_ind[i])} represents the indices <span class="math-inline">\(u,v\)</span> for the optimal assignment.
    \item This function correctly handles both square and rectangular cost matrices. For a rectangular matrix, it finds an optimal assignment using all rows or all columns, depending on which dimension is smaller. The total cost of the assignment is given by \texttt{cost\_matrix[row\_ind, col\_ind].sum()}.
\end{itemize}

Using ready-made library functions such as \texttt{scipy.optimize.linear\_sum\_assignment} is the preferred approach since they are well-tested, performance-optimized, and avoid the need for complex custom implementations.

\paragraph{Result of applying the algorithm:}
The result of applying the Hungarian algorithm (or its library analogue) to the cost matrix $\Cuv$ is a set of optimal neuron pairs $(u_1, v_1), (u_2, v_2), \dots, (u_k, v_k)$, where $k = \min(M,N)$. Each pair $(u_i, v_i)$ represents matching neuron $u_i$ from Layer 1 with neuron $v_i$ from Layer 2, and the total cost $\sum_{i=1}^k C_{u_i v_i}$ is minimal. This set of pairs will be used to compute the final similarity metric between layers.

\subsection{Defining the final metric (similarity / distance measure) between layers}
\label{ssec:final_layer_metric_definition}

After applying the Hungarian algorithm (or similar assignment solution method as described in Subsection~\ref{ssec:hungarian_algorithm_application}), we obtain a set of optimal pairwise correspondences between neurons of the two compared layers. Let Layer 1 have $N_1$ neurons and Layer 2 have $N_2$ neurons. The algorithm finds $M = \min(N_1, N_2)$ optimal neuron pairs $(u_1,v_1), (u_2,v_2), \dots, (u_M,v_M)$, where $u_i$ is a neuron from one layer and $v_i$ is the matched neuron from the other layer. Each such pair $(u_i,v_i)$ has an associated cost (distance) $C_{u_i v_i}$ from the cost matrix $C$ constructed in Subsection~\ref{ssec:cost_matrix_construction}. Recall that $C_{u_i v_i}$ represents the estimated Jaccard distance between neurons $u_i$ and $v_i$, computed based on their MinHash sketches.

Based on these optimal pairs and their costs, we can define the final metric that quantitatively characterizes the distance between the two layers. Denote this metric as $\LayerDist$. It is calculated as the mean cost of the optimal assignment:
$$ \LayerDist = \frac{1}{M} \sum_{i=1}^{M} C_{u_i v_i} $$
where $M$ is the number of optimal neuron pairs found (equal to $\min(N_1, N_2)$), and $C_{u_i v_i}$ is the cost of matching the $i$-th optimal pair.

\paragraph{Interpretation of the obtained $\LayerDist$ value:}
\begin{itemize}
    \item \textbf{Value range:} Since each cost $C_{u_i v_i}$ (estimated Jaccard distance) is in the range $[0, 1]$, the metric $\LayerDist$ will also be in the range $[0, 1]$.
    \item \textbf{Value close to 0:} Indicates high functional similarity between the layers. This means neurons from one layer can be successfully matched to neurons from the other, with the matched neurons having very similar activation patterns on average (low Jaccard distance).
    \item \textbf{Value close to 1:} Indicates low functional similarity (high difference) between the layers. Even with optimal matching, the average distance between corresponding neurons is large.
\end{itemize}

This metric $\LayerDist$ represents the average pairwise distance between the functionally closest neurons of the two layers (according to the chosen assignment algorithm and metric $C_{uv}$).

\paragraph{Discussion and possible nuances:}
\begin{itemize}
    \item \textbf{Dependence on parameters:} It is important to remember that the value of $\LayerDist$ depends on all previous steps: the choice and size of the sample $\XS$ for generating activation signatures, the number of hash functions $K$ in MinHash, and the quality of the Jaccard distance approximation.
    \item \textbf{Unmatched neurons:} If the layer sizes $N_1$ and $N_2$ differ, then $\max(N_1, N_2) - \min(N_1, N_2)$ neurons from the larger layer remain unmatched. The current metric $\LayerDist$ only accounts for matched pairs. In some tasks, it may be necessary to introduce a penalty for unmatched neurons or use another normalization strategy if the absolute number of neurons and completeness of matching are critical factors. However, for assessing similarity based on the “best possible alignment,” the proposed formula is the standard approach.
    \item \textbf{Similarity measure:} If a similarity measure (rather than distance) is required, it can be defined as $1 - \LayerDist$.
\end{itemize}

The introduced metric $\LayerDist$ provides a single quantitative value for assessing the functional closeness of two neural layers. This enables objective evaluation of how similar one layer is to another, which can be used for model analysis, tracking changes during training, comparing different architectures or initializations.

\section{Canonical Representation and Similarity Metric Algorithm: Summary and Complexity Analysis}
\label{sec:algorithm_summary_complexity}

This section provides a step-by-step description of the entire algorithm proposed in this work, from preprocessing neural network parameters to computing the final similarity metric between its layers. Key tuning parameter formulas are presented for the main stages. The second part of the section is devoted to analyzing the algorithmic complexity of each stage.

\subsection{Algorithm Stages and Key Parameters}
\label{ssec:algorithm_steps_parameters}

The algorithm consists of four main stages: preprocessing network parameters with normalization, generating sampled activation signatures for each analyzed layer's neurons, constructing compact MinHash sketches for these signatures, and finally computing the similarity metric between two layers based on optimal neuron matching.

\subsubsection{Preliminary L2-normalization of parameters (for each analyzed network)}
\label{sssec:algo_l2_normalization}

This stage is performed once for each neural network whose layers will be analyzed.
\begin{enumerate}
    \item \textbf{For each hidden layer $k$ of the network (sequentially):}
    \begin{itemize}
        \item For each neuron $j$ (with parameters $W_{k,j}, b_{k,j}$):
        \begin{itemize}
            \item Compute the L2-norm of the weight vector: $\rho_{k,j} = ||W_{k,j}||_2$.
            \item If $\rho_{k,j} \neq 0$:
            \begin{itemize}
                \item Scaling factor: $c_{k,j} = \rho_{k,j}$.
                \item Normalized parameters: $W'_{k,j} = W_{k,j}/c_{k,j}$, $b'_{k,j} = b_{k,j}/c_{k,j}$.
            \end{itemize}
            \item If $\rho_{k,j} = 0$:
            \begin{itemize}
                \item $c_{k,j} = 1$, $W'_{k,j} = W_{k,j}$, $b'_{k,j} = b_{k,j}$.
            \end{itemize}
        \end{itemize}
        \item Compensate the scaling factors in the weights of the next layer $k+1$: $W'_{k+1} = W_{k+1} \cdot \operatorname{diag}(c_{k,1}, \dots, c_{k,N_k})$.
    \end{itemize}
    \item \textbf{For the output layer $L$:}
    \begin{itemize}
        \item The weights $W_L$ are adjusted considering the scaling factors $c_{L-1,j}$ from the penultimate layer $L-1$. The output layer parameters $(W'_L, b_L)$ themselves are not further normalized.
    \end{itemize}
\end{enumerate}

\subsubsection{Generating sampled neuron activation signatures \texorpdfstring{($\Sj$)}{(Sj)} (for each analyzed layer \texorpdfstring{$k$}{k})}
\label{sssec:algo_sampled_signatures}

\begin{enumerate}
    \item \textbf{Determining the sample size $\Nsamp$ (see Subsection~\ref{subsubsec:sample_complexity_bounds}):}
    Choose $\Nsamp$ satisfying the condition (for given accuracy $\varepsilon$ and overall confidence $1-\delta_{VC}$ for all $N_k$ neurons in a layer with input dimension $D_{in}^{(k)}$):
    $$ \Nsamp \geq \frac{1}{\varepsilon^2}\left((D_{in}^{(k)}+1) \log\left(\frac{2e\Nsamp}{D_{in}^{(k)}+1}\right) + \log\left(\frac{2N_k}{\delta_{VC}}\right)\right) $$
    This equation is solved numerically for $\Nsamp$.

    \item \textbf{Generating the sample $X_S = \{\xs{1}, \dots, \xs{\Nsamp}\}$ (see Subsection~\ref{subsec:sampling_strategies}):}
    Generate or select $\Nsamp$ input vectors $\xs{s} \in \Rspace{D_{in}^{(k)}}$ using one of the strategies (e.g., existing dataset sampling, random sampling, structured sampling such as LHS or Sobol).

    \item \textbf{Computing sampled activation signatures $\Sj$ and the matrix $M_{act}$ (see Subsections~\ref{subsec:definition_SAR_Sj} and~\ref{subsec:activation_matrix_formation}):}
    For each neuron $j$ in layer $k$ (with normalized parameters $W'_{k,j}, b'_{k,j}$) and each sample $\xs{s} \in X_S$:
    $$ s_{j,s} = \indicate{W'_{k,j} \cdot \xs{s} + b'_{k,j} > 0} $$
    Form a binary matrix $M_{act}$ of size $N_k \times \Nsamp$, where $(M_{act})_{j,s} = s_{j,s}$.

    \item \textbf{Neuron filtering (see Subsection~\ref{subsec:primary_analysis_activation_matrix}):}
    \begin{itemize}
        \item Identify and filter out inactive (“dead”) neurons (where $\sum_s s_{j,s} = 0$).
        \item Identify and filter out always active neurons (where $\sum_s s_{j,s} = \Nsamp$).
        \item Group neurons with identical non-trivial signatures $\Sj$; select one representative for each group.
        \item Let $N_{uf,k}$ be the number of unique, non-trivially activating neurons in layer $k$ after filtering.
    \end{itemize}
\end{enumerate}

\subsubsection{Constructing MinHash sketches \texorpdfstring{($H(\Sj)$)}{(H(Sj))} (for each unique, non-trivial neuron \texorpdfstring{$j$}{j})}
\label{sssec:algo_minhash}

\begin{enumerate}
    \item \textbf{Determining the MinHash sketch length $K$ (see Subsection~\ref{sssec:minhash_parameter_estimation_reliability}):}
    Choose $K$ satisfying the condition (for a given error probability $\alpha_{MH}$ when distinguishing Jaccard distances differing by at least $\delta_J$):
    $$ K \geq \frac{2\ln(2/\alpha_{MH})}{\delta_J^2} $$

    \item \textbf{Computing MinHash sketches $H(\Sj)$ (see Subsection~\ref{ssec:minhash_function}):}
    For each sampled activation signature $\Sj$ (represented as the set of active sample indices $S_j^* = \{s \mid s_{j,s}=1\}$) and for each of the $K$ independent hash functions (or random permutations) $\pi_t$:
    $$ h_t(\Sj) = \min_{s \in S_j^*} \pi_t(s) $$
    The final MinHash sketch for neuron $j$ is the vector $H(\Sj) = [h_1(\Sj), \dots, h_K(\Sj)]$.
\end{enumerate}

\subsubsection{Computing the similarity metric between two layers \texorpdfstring{($\text{LayerDistance}$)}{(LayerDistance)} (Layer 1 and Layer 2)}
\label{sssec:algo_layer_distance}

Let Layer 1 have $N_{uf,1}$ unique neurons and Layer 2 have $N_{uf,2}$ unique neurons, for which MinHash sketches have been constructed.

\begin{enumerate}
    \item \textbf{Constructing the cost matrix $C$ (see Subsection~\ref{ssec:cost_matrix_construction}):}
    The matrix $C$ has size $N_{uf,1} \times N_{uf,2}$. Element $\Cuv$ (denoted here as $C_{uv}$ for indexing) equals the estimated Jaccard distance between neuron $u$ from Layer 1 and neuron $v$ from Layer 2:
    $$ C_{uv} = \rho(H(S_u), H(S_v)) = \frac{1}{K} \sum_{t=1}^{K} \indicate{h_t(S_u) \neq h_t(S_v)} $$

    \item \textbf{Finding the optimal matching (see Subsection~\ref{ssec:hungarian_algorithm_application}):}
    The Hungarian algorithm (or its analogue, e.g. \texttt{scipy.optimize.linear\_sum\_assignment}) is applied to the matrix $C$ to find $M_{opt} = \min(N_{uf,1}, N_{uf,2})$ optimal neuron pairs $(u_i, v_i)$ with corresponding costs $C_{u_i v_i}$, minimizing the total cost $\sum C_{u_i v_i}$.

    \item \textbf{Computing $\text{LayerDistance}$ (see Subsection~\ref{ssec:final_layer_metric_definition}):}
    $$ \text{LayerDistance} = \frac{1}{M_{opt}} \sum_{i=1}^{M_{opt}} C_{u_i v_i} $$
\end{enumerate}

\subsection{Algorithmic Complexity Analysis}
\label{ssec:complexity_analysis}

We estimate the computational complexity of each stage of the algorithm. Notation:
\begin{itemize}
    \item $L$: total number of layers in the network.
    \item $N_k$: number of neurons in layer $k$.
    \item $D_{in}^{(k)}$: input dimension of layer $k$.
    \item $\Nsamp$: sample size $X_S$.
    \item $N_{uf,k}$: number of unique, non-trivial neurons in layer $k$.
    \item $K$: MinHash sketch length.
    \item $\bar{w}_k$: average number of active samples per neuron in layer $k$. $0 \le \bar{w}_k \le \Nsamp$.
    \item When comparing two layers, $N_a = N_{uf,1}$ and $N_b = N_{uf,2}$.
\end{itemize}

\subsubsection{Complexity of L2-normalization stage (per network)}
\label{sssec:complexity_l2_norm}

\begin{itemize}
    \item For a single layer $k$:
    \begin{itemize}
        \item Normalizing $N_k$ neurons: $N_k \cdot \mathcal{O}(D_{in}^{(k)})$.
        \item Compensating in the next layer $k+1$ (with $N_{k+1}$ neurons): $\mathcal{O}(N_{k+1} \cdot N_k)$.
    \end{itemize}
    \item Overall for all $L-1$ hidden layers (assuming $N_k \approx N$ and $D_{in}^{(k)} \approx D$ for simplicity): approximately $\mathcal{O}(L \cdot (N D + N^2))$. This operation is performed once per network.
\end{itemize}

\subsubsection{Complexity of sampled activation signature computation stage \texorpdfstring{$\Sj$}{Sj} (per layer \texorpdfstring{$k$}{k})}
\label{sssec:complexity_sampled_signatures}

\begin{itemize}
    \item \textbf{Determining $\Nsamp$}: Solving the transcendental equation typically requires several iterations; this complexity is negligible or constant.
    \item \textbf{Generating $X_S$}:
    \begin{itemize}
        \item From existing data / random sampling: $\mathcal{O}(\Nsamp \cdot D_{in}^{(k)})$.
        \item Structured sampling: may vary, but often close to $\mathcal{O}(\Nsamp \cdot D_{in}^{(k)})$.
    \end{itemize}
    \item \textbf{Computing $\Sj$ (matrix $M_{act}$)}:
    \begin{itemize}
        \item Total: $\mathcal{O}(N_k \cdot \Nsamp \cdot D_{in}^{(k)})$.
    \end{itemize}
    \item \textbf{Neuron filtering}:
    \begin{itemize}
        \item Counting activations: $\mathcal{O}(N_k \cdot \Nsamp)$.
        \item Grouping identical signatures (with row hashing): $\mathcal{O}(N_k \cdot \Nsamp + N_k \log N_k)$. Dominated by $\mathcal{O}(N_k \cdot \Nsamp)$.
    \end{itemize}
\end{itemize}

\subsubsection{Complexity of MinHash sketch construction stage \texorpdfstring{$H(\Sj)$}{H(Sj)} (per layer \texorpdfstring{$k$}{k})}
\label{sssec:complexity_minhash}

\begin{itemize}
    \item \textbf{Determining $K$}: $\mathcal{O}(1)$.
    \item \textbf{Computing $H(\Sj)$ for $N_{uf,k}$ neurons}:
    \begin{itemize}
        \item For a single neuron $j$ with $w_j$ active samples: $\mathcal{O}(K \cdot w_j)$.
        \item Total for $N_{uf,k}$ neurons: $\mathcal{O}(N_{uf,k} \cdot K \cdot \bar{w}_k)$.
        \item In the worst case $\bar{w}_k = \Nsamp$: $\mathcal{O}(N_{uf,k} \cdot K \cdot \Nsamp)$.
    \end{itemize}
\end{itemize}

\subsubsection{Complexity of \texorpdfstring{$\text{LayerDistance}$}{LayerDistance} computation stage (for comparing two layers with \texorpdfstring{$N_a$}{Na} and \texorpdfstring{$N_b$}{Nb} unique neurons)}
\label{sssec:complexity_layer_distance}

\begin{itemize}
    \item \textbf{Constructing the cost matrix $C$ (size $N_a \times N_b$)}:
    \begin{itemize}
        \item Total: $\mathcal{O}(N_a \cdot N_b \cdot K)$.
    \end{itemize}
    \item \textbf{Finding the optimal matching (Hungarian algorithm)}:
    \begin{itemize}
        \item Complexity for an $N_a \times N_b$ matrix: $\mathcal{O}(\min(N_a, N_b)^2 \cdot \max(N_a, N_b))$. If $N_a \approx N_b \approx N_{uf}$, then $\mathcal{O}(N_{uf}^3)$.
    \end{itemize}
    \item \textbf{Computing $\text{LayerDistance}$}:
    \begin{itemize}
        \item Total: $\mathcal{O}(\min(N_a, N_b))$. Negligible.
    \end{itemize}
\end{itemize}

\textbf{Overall summary of dominant stages by complexity:}
\begin{itemize}
    \item \textbf{Initialization (per layer):} Generating sampled signatures $\Sj$: $\mathcal{O}(N_k \cdot \Nsamp \cdot D_{in}^{(k)})$.
    \item \textbf{Initialization (per layer):} Constructing MinHash sketches: $\mathcal{O}(N_{uf,k} \cdot K \cdot \bar{w}_k)$ (worst case $\mathcal{O}(N_{uf,k} \cdot K \cdot \Nsamp)$).
    \item \textbf{Layer comparison:} Constructing the cost matrix: $\mathcal{O}(N_a \cdot N_b \cdot K)$.
    \item \textbf{Layer comparison:} Hungarian algorithm: $\mathcal{O}(\min(N_a,N_b)^2 \cdot \max(N_a,N_b))$.
\end{itemize}

From the analysis, it is evident that the choice of parameters $\Nsamp$ and $K$, as well as the network characteristics (number of neurons $N_k$, input dimension $D_{in}^{(k)}$, and activation density $\bar{w}_k$), significantly affect the overall computational cost of the proposed approach. Optimizing these parameters and using efficient algorithm implementations are key for practical application to large neural networks.

The complexity analysis presented above allows us to conclude which stages contribute most to the overall computational cost of the proposed method. \textbf{The main computational bottlenecks are the stages related to processing each layer to obtain its characteristics before comparison, namely:}

\begin{enumerate}
    \item \textbf{Generating sampled activation signatures $\Sj$:} This stage has complexity $\mathcal{O}(N_k \cdot \Nsamp \cdot D_{in}^{(k)})$. The high cost is due to the need to compute the response of each of the $N_k$ neurons to each of the $\Nsamp$ samples, with each such computation involving operations with an input vector of dimension $D_{in}^{(k)}$. Considering that $\Nsamp$ theoretically needs to be large for representativeness (e.g., on the order of $10^6-10^7$, as suggested in Subsection~\ref{subsubsec:sample_complexity_bounds}), and that $N_k$ and $D_{in}^{(k)}$ can be hundreds or thousands, this stage is extremely resource-intensive.

    \item \textbf{Constructing MinHash sketches $H(\Sj)$:} This stage has complexity $\mathcal{O}(N_{uf,k} \cdot K \cdot \bar{w}_k)$, where $N_{uf,k}$ is the number of unique neurons, $K$ is the MinHash sketch length, and $\bar{w}_k$ is the average number of active samples per neuron. If neuron activations are not highly sparse (i.e., $\bar{w}_k$ is a significant fraction of $\Nsamp$), this stage also becomes very computationally intensive. In the extreme case where $\bar{w}_k \approx \Nsamp$, its complexity $\mathcal{O}(N_{uf,k} \cdot K \cdot \Nsamp)$ may be comparable to or even exceed the complexity of generating the original signatures $\Sj$, especially if $K$ (which can be on the order of $10^2-10^3$) exceeds $D_{in}^{(k)}$. However, with high activation sparsity ($\bar{w}_k \ll \Nsamp$), the cost of this stage is significantly reduced.
\end{enumerate}

Which of these two initialization stages becomes absolutely dominant depends on the specific parameter values, particularly the average activation density $\bar{w}_k$ and the relative sizes of $K$ and $D_{in}^{(k)}$.

In comparison to these two layer-wise “initialization” stages, subsequent steps related directly to comparing two prepared layers (constructing the cost matrix $\mathcal{O}(N_a \cdot N_b \cdot K)$ and applying the Hungarian algorithm $\mathcal{O}(\min(N_a,N_b)^2 \cdot \max(N_a,N_b))$) are generally less costly, as their complexity does not directly depend on the huge sample size $\Nsamp$ or the original data dimension $D_{in}^{(k)}$.

\textbf{Thus, the key factor determining the overall performance of the proposed approach is the efficiency of processing the large data volume ($\Nsamp$ samples) for characterizing each neuron in the initial stages.} Optimizing these operations, as well as reasonable choice of $\Nsamp$ and $K$ parameters given available computational resources, are of primary importance for practical applicability of the method to analyzing modern deep neural networks.

\section{Analysis of the Properties of the Proposed Distance Measure}
\label{sec:metric_properties_analysis}

This section examines the formal properties of the distance measure $\rho(A,B) = \frac{1}{K}\sum_{t=1}^{K} \mathbb{1}_{\{h_t(A) \neq h_t(B)\}}$, used to estimate the true Jaccard distance $d_J(A,B)$ between sampled activation signatures of neurons. While $d_J(A,B)$ is a proper metric, the properties of its estimator $\rho(A,B)$ require careful consideration. This measure $\rho$ was introduced in previous sections as an efficient similarity estimator.

\subsection{Non-negativity}
\label{subsec:non_negativity_rho}

The non-negativity property for $\rho(A,B)$ holds trivially.

\textbf{Proposition 9.1.} For any sets $A, B$, $0 \leq \rho(A,B) \leq 1$.

\textbf{Proof.} Each indicator function $\mathbb{1}_{\{h_t(A) \neq h_t(B)\}}$ takes values in $\{0,1\}$. Therefore, their arithmetic mean also lies within $[0,1]$. $\square$

\subsection{Identity}
\label{subsec:identity_rho}

The identity property for the estimator $\rho(A,B)$ requires more careful analysis due to the probabilistic nature of hashing.

\textbf{Proposition 9.2.} Let $A$ and $B$ be sets from the universe, and let $h_1, \ldots, h_K$ be independent MinHash functions. Then:
\begin{enumerate}
    \item If $A = B$, then $\rho(A,B) = 0$ (deterministically).
    \item If $A \neq B$, then $\Pr[\rho(A,B) = 0 \mid A \neq B] = J(A,B)^K$, where $J(A,B)$ is the Jaccard index.
\end{enumerate}

\textbf{Proof.}
\begin{enumerate}
    \item If $A = B$, then for any MinHash function $h_t$ (based on permutation $\pi_t$ or universal hash function $H_t$), we have $h_t(A) = h_t(B)$. Therefore, all indicators $\mathbb{1}_{\{h_t(A) \neq h_t(B)\}}$ equal zero, and thus $\rho(A,B) = 0$.
    \item If $A \neq B$, then $\rho(A,B) = 0$ if and only if $h_t(A) = h_t(B)$ for all $t = 1, \ldots, K$. Since the hash functions are independent, and $\Pr[h_t(A) = h_t(B)] = J(A,B)$ for each function (by the MinHash property, see Subsection~\ref{sssec:minhash_jaccard_property}), we get:
    $$\Pr[\rho(A,B) = 0 \mid A \neq B] = \prod_{t=1}^{K} \Pr[h_t(A) = h_t(B)] = J(A,B)^K$$
\end{enumerate}
$\square$

\textbf{Corollary.} For distinct sets $A \neq B$ with $J(A,B) < 1$, the probability of false equality $\rho(A,B) = 0$ is exponentially small for sufficiently large $K$. For example, if $J(A,B) = 0.9$ and $K = 100$: $\Pr[\rho(A,B) = 0 \mid A \neq B] = 0.9^{100} \approx 2.7 \times 10^{-5}$.

Thus, if $A=B$, then $\rho(A,B)=0$ deterministically. For distinct $A,B$ (where $J(A,B)<1$), the probability of false equality $\rho(A,B)=0$ is $J(A,B)^K$. Therefore, the full \textbf{identity property ($\rho(A,B)=0 \iff A=B$) holds for the MinHash estimator $\rho(A,B)$ with probability tending to 1 for distinct $A,B$ (where $J(A,B)<1$) as $K$ increases.}

\subsection{Symmetry}
\label{subsec:symmetry_rho}

\textbf{Proposition 9.3.} For any sets $A, B$, $\rho(A,B) = \rho(B,A)$.

\textbf{Proof.} This follows directly from the symmetry of the condition $h_t(A) \neq h_t(B)$ (and the symmetry of the Jaccard distance $d_J(A,B)$, which it estimates). $\square$

\subsection{Triangle inequality (satisfaction or violation)}
\label{subsec:triangle_inequality_rho}

For a proper metric, the triangle inequality must hold. In the case of the MinHash estimator $\rho(A,B)$, this property holds for the expectation but may be violated for specific realizations.

\textbf{Proposition 9.4.} For the expectation of the MinHash estimator $\rho$, the triangle inequality holds:
$$\mathbb{E}[\rho(A,C)] \leq \mathbb{E}[\rho(A,B)] + \mathbb{E}[\rho(B,C)]$$

\textbf{Proof.} Since $\mathbb{E}[\rho(A,B)] = d_J(A,B)$ (the Jaccard distance, see Subsection~\ref{ssec:jaccard_distance_estimation_minhash}), and the true Jaccard distance $d_J$ satisfies the triangle inequality ($d_J(A,C) \leq d_J(A,B) + d_J(B,C)$), the proposition follows directly. $\square$

\textbf{Note.} For a specific realization $\rho(A,B)$ with a fixed set of $K$ hash functions, the triangle inequality $\rho(A,C) \leq \rho(A,B) + \rho(B,C)$ may be violated due to statistical fluctuations in the estimator. The probability of such violation decreases as $K$ increases, since the estimator $\rho$ becomes more concentrated around its expectation. This means that $\rho(A,B)$ does not always satisfy the triangle inequality for specific values, although its expectation does.

\subsection{Sensitivity analysis of the distance measure}
\label{subsec:sensitivity_analysis_rho}

The main parameters affecting the properties of the MinHash-based distance estimate $\rho(A,B)$ are the number of hash functions $K$ and the characteristics of the original data, in particular, the sample size $\Nsamp$ on which the activation signatures $\Sj$ (represented as sets $A, B, \dots$) were constructed.

\paragraph{Effect of the number of hash functions $K$:}
The parameter $K$ directly influences the accuracy and reliability of the estimate $\rho(A,B)$ and its associated properties:
\begin{itemize}
    \item \textbf{Order preservation:} As shown in the theorem from Subsection~\ref{sssec:order_preservation_theorem_statement}, the probability of correctly determining the relative order of two different Jaccard distances based on their $\rho$ estimates increases as $1 - 2\exp(-K\delta^2/2)$, i.e. exponentially with $K$.
    \item \textbf{Accuracy of estimating $d_J(A,B)$:} The variance of the estimate $\rho(A,B)$ of the Jaccard distance $d_J(A,B)$ decreases as $O(1/K)$ (more precisely, $\text{Var}[\rho(A,B)] = \frac{d_J(A,B)(1-d_J(A,B))}{K}$). Accordingly, the standard deviation decreases as $O(1/\sqrt{K})$.
    \item \textbf{Identity property:} The probability of a false equality ($\rho(A,B)=0$ when $A \neq B$) decreases exponentially with increasing $K$ as $J(A,B)^K$.
    \item \textbf{Computational complexity:} The time to build a MinHash sketch and to compare it with another sketch increases linearly with $K$.
\end{itemize}
Thus, increasing $K$ improves the statistical properties of $\rho(A,B)$, bringing it closer to the true metric $d_J(A,B)$, but increases computational costs.

\paragraph{Effect of data characteristics (via $\Nsamp$):}
The parameter $\Nsamp$ (the sample size on which the signatures $\Sj$ are formed) does not directly affect the mathematical properties of the estimation procedure $\rho$ for already given sets $A$ and $B$. However, it is critically important for how well these sets $A$ and $B$ (obtained from $\Sj$) represent the true functional characteristics of neurons:
\begin{itemize}
    \item \textbf{Accuracy of activation region approximation:} The quality of representing the continuous activation regions $\ARj$ with discrete signatures $\Sj$ (and corresponding sets) directly depends on $\Nsamp$, as discussed in the context of VC dimension (Subsection~\ref{subsubsec:vc_interpretation_practical_choice}).
    \item \textbf{Resolution of the original signatures:} The minimum possible nonzero Jaccard distance between two different signatures $\Sj$ is limited to the order of $1/\Nsamp$. If $\Nsamp$ is too small, many functionally different neurons may obtain identical or very similar signatures $\Sj$, making them indistinguishable in subsequent stages, including $\rho$ estimation.
\end{itemize}
Therefore, an adequate choice of $\Nsamp$ is a necessary condition for the subsequent analysis using MinHash and $\rho(A,B)$ to be meaningful.

\paragraph{Final conclusion on the properties of $\rho(A,B)$:}
In conclusion, the Jaccard distance estimate $\rho(A,B)$ based on MinHash is non-negative and symmetric. The property $\rho(A,A)=0$ always holds. However, other key metric axioms – identity for distinct objects and the triangle inequality – hold for $\rho(A,B)$ in a probabilistic sense, and their reliability increases with the number of hash functions $K$. This makes $\rho(A,B)$ a practically valuable and efficient estimate of the true Jaccard distance for applied tasks.

\section{Experimental Evaluation and Results}

\subsection{Objectives of the experiments}
\label{subsec:experimental_aims}

The experimental part of this work pursued several key objectives aimed at empirically verifying and demonstrating the capabilities of the proposed approach for canonical representation and similarity metric computation of neural networks.

Firstly, the main objective was to \textbf{demonstrate the practical applicability of the developed algorithm and the introduced \LayerDist\ metric}. To this end, a quantitative comparison was conducted to assess the functional similarity of hidden layers of two neural networks. Importantly, these networks had identical architectures and were trained to solve the same binary classification task. This scenario is a typical example where, despite having the same task formulation and architecture, random factors (such as weight initialization) can lead to the formation of different internal representations by neurons. The ability of the \LayerDist\ metric to objectively assess the degree of similarity of such representations is a central aspect of this evaluation.

Secondly, since the proposed algorithm uses MinHash signatures to approximate neuron activation regions for computational efficiency, an important goal was to \textbf{assess the accuracy of this approximation}. Specifically, the accuracy of estimating Jaccard distances between sample activation signatures of neurons using MinHash sketches with a given number of hash functions ($K=512$) was investigated. This assessment was performed by comparing the results obtained via MinHash with those computed from exact (full) activation signatures, providing insight into the reliability and adequacy of hashing in this context.

Thirdly, the objective was to \textbf{visually compare and analyze the internal representations} learned by the compared neural networks. By visualizing the decision boundaries of individual hidden layer neurons, it was intended to clearly demonstrate how different weight configurations lead to different ways of partitioning the input space, even if the overall task is solved similarly. This qualitative observation was meant to further highlight the problem of representational ambiguity and the relevance of developing objective similarity metrics.

Achieving these objectives enables drawing conclusions about the practical value and characteristics of the developed method.

\subsection{Methodology}
\label{subsec:methodology}

This section describes the methodology of the experiments, including data preparation, neural network configuration, training procedures, and subsequent analysis for computing the similarity metric.

\subsubsection{Parameters and sampling for activation analysis \texorpdfstring{($X_S$)}{(XS)}}
\label{ssubsec:methodology_sampling}

To characterize neuron activation regions, an input data sample $\XS$ was used. The theoretical minimum size of this sample, $\Nsamp$, was estimated using the formula derived from Vapnik–Chervonenkis theory (as described in Subsection~\ref{subsubsec:sample_complexity_bounds}). For the analyzed hidden layer, the parameters were:

\begin{itemize}
    \item Number of neurons ($N_k$): 32
    \item Input space dimensionality ($D_{in}^{(k)}$): 2
    \item Desired probability estimation accuracy ($\varepsilon$): 0.05
    \item Overall confidence level ($1-\delta$): 0.99 (i.e., $\delta = 0.01$)
\end{itemize}

The calculation yielded a minimum required sample size of $\Nsamp \ge 15823$.

For practical purposes, a sample $\XS$ of \textbf{16,000 points} was generated. These points were created using the \textbf{Latin Hypercube Sampling (LHS)} method (via \texttt{scipy.stats.qmc.LatinHypercube}) to ensure good and uniform coverage of the two-dimensional input space. The generated points from the unit square $[0, 1]^2$ were linearly scaled to the target range $[-10, 10] \times [-10, 10]$. This sample was used in subsequent analysis stages.

\subsubsection{Neural network configuration and training}
\label{ssubsec:methodology_nn_training}

The experiment used \textbf{two neural networks} with identical architectures, implemented using TensorFlow (Keras API). Each network had the following structure:

\begin{itemize}
    \item \textbf{Input layer:} Received 2 features (coordinates $x_1, x_2$ from the sample $\XS$).
    \item \textbf{Hidden layer:} Consisted of 32 neurons with \textbf{ReLU} activation functions. A key feature of this layer was the application of a unit norm constraint on the incoming weights of each neuron: the sum of squares of weights was enforced to equal 1 using \texttt{tf.keras.constraints.UnitNorm(axis=0)}. This ensured L2 normalization of hidden layer weights during training, consistent with the theoretical part of this work (Subsection~\ref{subsec:l2_normalization}).
    \item \textbf{Output layer:} Consisted of 1 neuron with a \textbf{sigmoid} activation function for binary classification.
\end{itemize}

The \textbf{task} for the neural networks was to classify points from the sample $\XS$ relative to an ellipse defined by the equation $\frac{x_1^2}{9} + \frac{x_2^2}{16} = 1$. Training labels were formed as follows: points inside the ellipse ($\frac{x_1^2}{9} + \frac{x_2^2}{16} - 1 < 0$) belonged to class 1, others to class 0.

Prior to training, the data was split into training and test sets with an 80/20 ratio. Each network was trained for \textbf{20 epochs} using:

\begin{itemize}
    \item Optimizer: \textbf{Adam}.
    \item Loss function: \textbf{Binary Crossentropy}.
    \item Evaluation metric: \textbf{Accuracy}.
    \item Batch size: 32.
\end{itemize}

After 20 epochs, both networks achieved similar performance on the test set: accuracy of approximately \textbf{0.99}, and loss around \textbf{0.02}. During training, model weights were saved at each epoch using \texttt{tf.keras.callbacks.ModelCheckpoint}. For final comparison, models obtained after 20 epochs were used.

\subsubsection{Obtaining neuron characteristics}
\label{ssubsec:methodology_neuron_features}

For each of the two trained neural networks, the following hidden layer neuron characteristics were sequentially obtained:
\begin{enumerate}
    \item \textbf{Sample activation signature matrix ($M_{act}$):} Using the sample $\XS$ (16,000 points) and the weights and biases of each trained network's hidden layer, a matrix $M_{act}$ of size $32 \times 16000$ was computed. Each element $(j,s)$ of this matrix equals 1 if neuron $j$ was active for the $s$-th sample from $\XS$, and 0 otherwise.
    \item \textbf{MinHash signatures ($H(\Sj)$):} The obtained activation matrices $M_{act}$ were used to generate MinHash signatures (sketches) for each of the 32 neurons. The \texttt{datasketch.MinHash} library was used with $K=512$ independent hash functions (parameters \texttt{num\_perm=512} and \texttt{seed=42} for reproducibility). As a result, each layer yielded a MinHash signature matrix of size $32 \times 512$.
\end{enumerate}

\subsubsection{Layer comparison procedure and computation of the \texorpdfstring{$\text{LayerDistance}$}{LayerDistance} metric}
\label{ssubsec:methodology_comparison_metric}

Comparison of the hidden layers of the two trained neural networks included the following steps:
\begin{enumerate}
    \item \textbf{Computing the cost matrix ($C$):} A matrix $C$ of size $32 \times 32$ was built. Each element $C_{uv}$ represents the estimated Jaccard distance between the MinHash signature of neuron $u$ from the first network and neuron $v$ from the second network, computed as the fraction of non-matching hash values ($\frac{1}{K} \sum_{t=1}^{K} \mathbb{I}(h_t(S_u) \neq h_t(S_v))$). (Note: the command $\mathbb{I}$ is defined as \texttt{\textbackslash indicate} in your document, assumed here to be correct.)
    \item \textbf{Finding optimal matching:} The Hungarian algorithm (implemented via \texttt{scipy.optimize.linear\_sum\_assignment}) was applied to the cost matrix $C$ to find the optimal one-to-one matching between the 32 neurons of the first network and the 32 neurons of the second network, minimizing total matching cost.
    \item \textbf{Computing the $\text{LayerDistance}$ metric:} The final similarity (distance) metric $\text{LayerDistance}$ between the two layers was calculated as the mean of the costs (estimated Jaccard distances) for the optimally matched neuron pairs.
\end{enumerate}

\subsubsection{Validation of MinHash approximation accuracy}
\label{ssubsec:methodology_minhash_validation}

To assess the quality of Jaccard distance approximation using MinHash signatures, the following additional steps were performed:
\begin{enumerate}
    \item The exact Jaccard distance matrix between all neuron pairs from the two networks was computed based on their full sample activation signatures $M_{act}$.
    \item Optimal matching was also performed using this "ideal" distance matrix.
    \item The matrix of absolute differences between the exact Jaccard distances and their MinHash estimates was computed.
    \item Final approximation error metrics were calculated: mean absolute error (MAE) and root mean squared error (RMSE).
\end{enumerate}

\subsection{Presentation, analysis, and interpretation of experimental results}
\label{subsec:results_presentation}

This section presents the results of the experimental evaluation of the proposed method. The analysis includes both qualitative visual assessment of the representations learned by the neural networks and quantitative metrics characterizing layer similarity and approximation accuracy.

\subsubsection{Visual analysis of learned representations}
\label{ssubsec:results_visual_analysis}

\begin{figure}[H]
    \centering

    \begin{subfigure}[b]{0.49\textwidth}
        \centering
        \includegraphics[width=\textwidth]{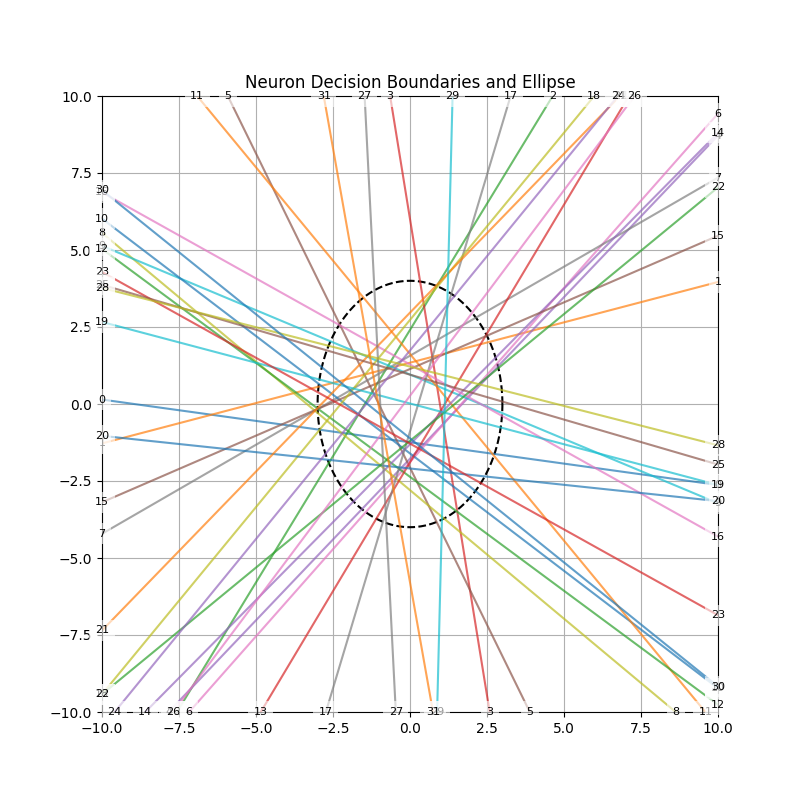}
        \caption{Network 1}
        \label{fig:weights_net1_sub}
    \end{subfigure}
    \hfill
    \begin{subfigure}[b]{0.49\textwidth}
        \centering
        \includegraphics[width=\textwidth]{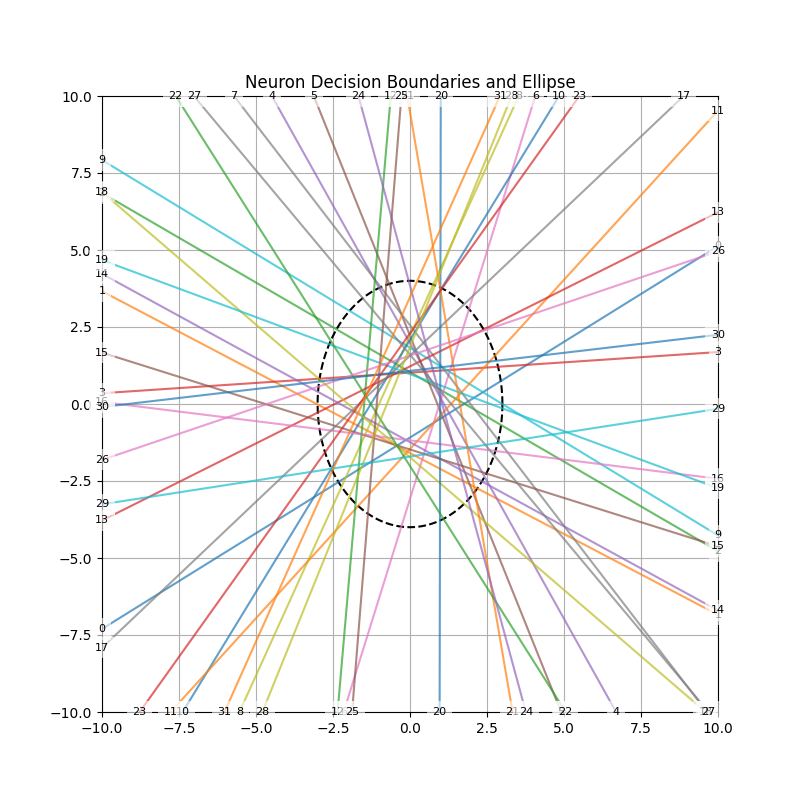}
        \caption{Network 2}
        \label{fig:weights_net2_sub}
    \end{subfigure}

    \caption{\textit{Visualization of hidden layer neuron decision boundaries for the two trained networks: (a) Network 1 and (b) Network 2.}}
    \label{fig:weights_both_nets}
\end{figure}

For qualitative analysis of the internal representations formed by the two trained neural networks (hereafter Network 1 and Network 2), the decision boundaries of their hidden layer neurons were visualized. Figures~\ref{fig:weights_net1_sub} and \ref{fig:weights_net2_sub} show these boundaries for Network 1 and Network 2, respectively, overlaid on the target ellipse boundary.

Visual comparison of these two figures clearly demonstrates that despite training on the same task and achieving similar accuracy (approximately 0.99 for both networks, as stated in the methodology), the internal weight configurations and, consequently, the geometry of individual neuron decision boundaries differ significantly. The orientation and position of lines for neurons with the same indices in the two networks do not match. This is a classic example of the representation ambiguity problem in neural networks, where equivalent or similar functions can be implemented by many different parameter sets. This observation underscores the need for metrics capable of assessing functional similarity independent of specific parameterizations.

\subsubsection{Quantitative assessment of layer similarity using \LayerDist}
\label{ssubsec:results_layerdistance}

To quantitatively assess the functional similarity between the hidden layers (each consisting of 32 neurons) of Network 1 and Network 2, the proposed \LayerDist\ metric was computed. The calculation was based on the MinHash signatures ($K=512$) of the neurons and the application of the Hungarian algorithm to find the optimal pairwise matching, as described in the methodology section.

The obtained \LayerDist\ metric value between the hidden layers of the two networks was \textbf{0.1696}.

Since the \LayerDist\ metric takes values in the range from 0 (complete identity of matched neurons by their activation signatures) to 1 (complete dissimilarity), the obtained value of 0.1696 indicates a relatively high degree of functional similarity between the two analyzed layers. This shows that despite visually different weight configurations (as shown in Figures~\ref{fig:weights_net1_sub} and \ref{fig:weights_net2_sub}), at the level of sampled activation patterns, neurons of one layer can be successfully matched with functionally similar neurons of the other layer.

\subsubsection{Accuracy assessment of the MinHash approximation}
\label{ssubsec:results_minhash_accuracy}

An important aspect of the proposed method is the use of MinHash signatures for efficient estimation of Jaccard distances between sampled neuron activation signatures. To verify the adequacy of this approximation under the conducted experiment (with $K=512$ hash functions), error metrics were calculated in comparison with the exact Jaccard distances computed from the full activation signatures ($32 \times 16000$).

Comparison of the estimated Jaccard distance matrix (obtained using MinHash) with the exact Jaccard distance matrix for all $32 \times 32 = 1024$ neuron pairs between the two networks yielded the following results:
\begin{itemize}
    \item MAE: \textbf{0.013560}
    \item RMSE: \textbf{0.017265}
\end{itemize}
These low error values indicate that the MinHash approximation with $K=512$ provides a sufficiently accurate representation of the original Jaccard distances. The average deviation of the estimate from the true value is around 0.01–0.02.

For additional context, the \LayerDist\ value computed based on the \textit{exact} Jaccard distances (using optimal matching with these exact distances) was \textbf{0.1560}. This value is very close to the value obtained using MinHash (0.1696), further confirming the suitability of the MinHash approximation for calculating the final layer similarity metric in this experimental scenario.

\subsubsection{Summary of neuron matching results}
\label{ssubsec:results_matching_tables}

For a detailed analysis of the results of optimal pairwise neuron matching between the hidden layers of Network 1 and Network 2, summary tables were created.

Table~\ref{tab:minhash_matching} shows the matching results obtained based on MinHash signatures (with $K=512$). For each optimal pair of neurons found (neuron $u$ from Network 1 and neuron $v$ from Network 2), the corresponding estimated Jaccard distance is given.

\begin{table}[H]
\centering
\caption{\textit{Optimal neuron matching and Jaccard distances (MinHash estimate, $K=512$)}.}
\label{tab:minhash_matching}
\begin{tabular}{ccc}
\hline
Neuron (Network 1) & Neuron (Network 2) & Distance (MinHash) \\
\hline
0 & 16 & 0.01563 \\
1 & 26 & 0.03125 \\
2 & 10 & 0.02148 \\
3 & 24 & 0.05469 \\
4 & 3 & 0.39844 \\
5 & 21 & 0.22852 \\
6 & 29 & 0.34961 \\
7 & 12 & 0.43555 \\
8 & 18 & 0.08398 \\
9 & 2 & 0.08398 \\
10 & 22 & 0.31055 \\
11 & 27 & 0.00391 \\
12 & 1 & 0.10938 \\
13 & 6 & 0.14063 \\
14 & 11 & 0.03320 \\
15 & 30 & 0.13867 \\
16 & 9 & 0.04297 \\
17 & 28 & 0.09961 \\
18 & 23 & 0.04492 \\
19 & 4 & 0.83398 \\
20 & 15 & 0.08203 \\
21 & 31 & 0.18750 \\
22 & 0 & 0.12891 \\
23 & 14 & 0.00977 \\
24 & 17 & 0.10547 \\
25 & 19 & 0.05273 \\
26 & 13 & 0.33203 \\
27 & 25 & 0.07031 \\
28 & 5 & 0.75586 \\
29 & 20 & 0.01172 \\
30 & 7 & 0.27930 \\
31 & 8 & 0.28906 \\
\hline
\end{tabular}
\end{table}

Similarly, Table~\ref{tab:exact_jaccard_matching} presents the results of optimal neuron matching, but based on exact Jaccard distances computed from the full sampled activation signatures.

\begin{table}[H]
\centering
\caption{\textit{Optimal neuron matching and exact Jaccard distances.}}
\label{tab:exact_jaccard_matching}
\begin{tabular}{ccc}
\hline
Neuron (Network 1) & Neuron (Network 2) & Distance (exact) \\
\hline
0 & 16 & 0.01234 \\
1 & 26 & 0.03555 \\
2 & 10 & 0.02783 \\
3 & 24 & 0.05707 \\
4 & 29 & 0.31499 \\
5 & 21 & 0.23142 \\
6 & 3 & 0.42024 \\
7 & 12 & 0.44401 \\
8 & 18 & 0.07825 \\
9 & 2 & 0.08447 \\
10 & 22 & 0.28955 \\
11 & 27 & 0.00514 \\
12 & 1 & 0.09471 \\
13 & 6 & 0.15051 \\
14 & 11 & 0.04317 \\
15 & 30 & 0.16183 \\
16 & 9 & 0.04441 \\
17 & 28 & 0.10437 \\
18 & 23 & 0.05643 \\
19 & 5 & 0.81998 \\
20 & 15 & 0.09128 \\
21 & 31 & 0.19622 \\
22 & 0 & 0.11023 \\
23 & 14 & 0.00706 \\
24 & 17 & 0.11291 \\
25 & 19 & 0.04341 \\
26 & 13 & 0.33177 \\
27 & 25 & 0.06885 \\
28 & 4 & 0.82497 \\
29 & 20 & 0.01801 \\
30 & 7 & 0.27280 \\
31 & 8 & 0.30018 \\
\hline
\end{tabular}
\end{table}

Comparing these two tables makes it possible to evaluate not only the quantitative value of the final \LayerDist\ metric but also the stability of the neuron matching process itself when using the MinHash approximation. For example, one can analyze what percentage of neuron pairs $(u,v)$ coincide in both tables.

\subsubsection{Discussion of results}
\label{ssubsec:results_discussion}

The obtained results demonstrate that the proposed approach allows for quantitative assessment of functional similarity between neural network layers, which, despite being trained on the same task and achieving similar solution quality, may have different internal parameterizations. The \LayerDist\ metric yielded a specific value (0.1696) characterizing this similarity.

The high accuracy of the MinHash approximation ($K=512$) for estimating Jaccard distances in this experiment (MAE $\approx 0.0136$, RMSE $\approx 0.0173$) was confirmed, making the method computationally attractive compared to calculations on full activation signatures without significant loss of accuracy in the final layer similarity estimate. The visual analysis (Figures~\ref{fig:weights_net1_sub} and \ref{fig:weights_net2_sub}) confirms the presence of the representation ambiguity problem and illustrates the task that the proposed metric addresses. Analysis of Tables~\ref{tab:minhash_matching} and \ref{tab:exact_jaccard_matching} shows that although optimal pairs may differ slightly, the overall similarity pattern is preserved.

\section{Conclusion and Directions for Future Research}
\label{sec:conclusion_future_work}

\subsection{Conceptual results and contributions of this work}
\label{subsec:conceptual_results_contribution}

This work asserts and systematically develops an approach to neural networks as formal mathematical objects, taking their analysis beyond purely engineering and algorithmic practice. The key conceptual result is the establishment of the possibility of introducing a (pseudo)metric structure on the space of neural network layers with identical architectures. This is achieved through the development and theoretical justification of the functional similarity metric $\LayerDist$. Introducing such a metric is a fundamental contribution, as it paves the way for applying the rich toolkit of metric geometry and topology to the study of neural networks, creating prerequisites for a qualitatively new level of theoretical understanding that was previously hindered by the lack of adequate formal tools for comparison.

The theoretical and methodological basis of this achievement lies firstly in the formalization of the problem of representational ambiguity in neural networks, including the systematization of the understanding of symmetry groups (scaling and permutation transformations) in ReLU networks, and the proposal of a mathematically justified approach to eliminating scaling ambiguity via L2 normalization. Secondly, a new concept was proposed for characterizing neurons through the fundamental notion of the "activation region" ($\AR{j}$) as a geometric characteristic of their functional contribution, with a developed transition to its practical discrete approximation through "sampled activation signatures" ($\Sj$), which are robust to small parameter perturbations.

The central metric contribution is the $\LayerDist$ metric itself – a quantitative measure of functional similarity based on the average pairwise distance between optimally matched neurons, with a clear interpretation in the $[0,1]$ range. Analysis of its properties confirmed non-negativity, symmetry, and examined the conditions for identity and the triangle inequality, as well as its sensitivity to parameters. This not only allows for comparing neural layers but also lays the foundation for studying the structure of the neural network space itself.

The realization of these concepts was made possible thanks to a significant algorithmic contribution. The MinHash technique was adapted for efficient compressed representation of neuron characteristics with theoretically proven order-preserving bounds and practical recommendations for parameter selection. Formalizing the neuron matching problem as an assignment problem with subsequent integration of the Hungarian algorithm enabled finding optimal correspondences. The practical contribution of the work is confirmed by a detailed analysis of the computational efficiency of the proposed method and its successful experimental validation. High accuracy of the MinHash approximation and the ability of the metric to detect functional similarity between networks with different weight configurations were demonstrated.

Thus, this work marks a conceptual shift from analyzing specific weight values to analyzing the functional behavior of neurons and networks as a whole. It not only shows that functionally equivalent networks can have different parameterizations but also lays the foundation for developing a theory of canonical representations, opening up possibilities for objective analysis of learning dynamics, knowledge transfer, model merging, and the development of neural network interpretation methods. Overall, successful integration of theoretical concepts, algorithmic techniques, and practical considerations has been demonstrated, forming a comprehensive methodology for solving a wide range of neural network analysis tasks.

\subsection{Discussion of the strengths and limitations of the proposed approach and metric}
\label{subsec:strengths_limitations}

The approach to canonical representation and similarity metric definition proposed in this work has several significant advantages, but like any new method, it has certain limitations that should be taken into account.

\paragraph{Strengths of the proposed approach}

One of the key strengths of the developed method is its deep theoretical elaboration and mathematical rigor. Unlike many heuristic approaches, the proposed method relies on solid mathematical foundations: using Vapnik–Chervonenkis theory to justify the sample size $\XS$ provides statistical guarantees of the representativeness of the obtained neuron characteristics; applying locality-sensitive hashing (LSH) theory with proven metric preservation bounds of the MinHash approximation ensures reliable similarity estimation; and the geometric interpretation of neuron "activation regions" ($\AR{j}$) gives the method a clear physical meaning.

An important advantage is robustness to network parameter variations. The proposed approach, unlike traditional methods based on direct comparison of weight coefficients or simple activation statistics, demonstrates significantly greater robustness. Using activation regions as the basic functional characteristic of a neuron ensures continuous dependence on its parameters, and MinHash representations inherit this robustness. The resulting $\LayerDist$ metric reflects functional rather than parametric similarity, making it invariant to equivalent representation transformations (e.g., neuron permutations or symmetric scalings eliminated by preliminary L2 normalization).

Computational scalability is also a significant advantage. The developed method effectively addresses the problem of high computational complexity inherent in analyzing large neural networks. The transition from $O(N_k^2 \cdot \Nsamp)$ operations for direct pairwise comparison of full activation signatures to $O(N_k^2 \cdot K)$ operations using MinHash sketches for cost matrix construction significantly reduces computational costs at the comparison stage. The ability to adjust the parameter $K$ (number of hash functions) allows for flexible balancing between approximation accuracy and computation speed. Furthermore, the algorithm allows for efficient parallelization at the stages of forming sampled activation signatures and computing the cost matrix.

Finally, the method has clear practical applicability and interpretability. The $\LayerDist$ metric takes values in an intuitively understandable $[0,1]$ range, where 0 corresponds to complete functional match (at the level of MinHash signatures of matched neurons), and 1 to complete dissimilarity. The optimal neuron matching procedure using the Hungarian algorithm provides detailed information on the most probable structural correspondences between functional units of the compared layers. Qualitative assessment of differences in learned representations is also possible by visualizing neuron activation regions or decision boundaries.

\paragraph{Limitations of the proposed approach}

Despite the listed strengths, the current implementation of the method and the metric itself have several limitations regarding neural network architecture and the type of activation functions used. The method was developed and tested primarily for fully connected layers with ReLU activation functions. Its direct application to more complex architectures, such as convolutional, recurrent layers, or attention-based mechanisms, would require significant adaptation and modification of both the "activation region" concept and the normalization and comparison procedures. Similarly, using other nonlinear activation functions (e.g., sigmoid, tanh, GELU) may necessitate a revision of the theoretical foundations of the method, particularly the L2 normalization procedure and its impact on subsequent layers.

The quality and reliability of the results critically depend on the choice and representativeness of the sample $\XS$ used for neuron characterization. If the sample $\XS$ does not adequately cover important regions of the input space where the network exhibits different behaviors, the method may miss significant functional differences or, conversely, overestimate similarity. For tasks with high-dimensional inputs (e.g., image or text analysis), creating a truly representative sample $\XS$ of sufficient size ($\Nsamp$) becomes a challenging problem in itself. The method implicitly assumes that the distribution of points in $\XS$ adequately reflects the actual data distribution encountered by the neural network.

Although using MinHash significantly increases scalability at the comparison stage, computational resource requirements during initialization (obtaining signatures for each analyzed layer) remain high. Forming the full activation matrix $M_{act}$ of size $N_k \times \Nsamp$ requires substantial memory, especially for theoretically justified sample sizes $\Nsamp$ (on the order of $10^6 - 10^7$ elements or more). Computing activations for all $N_k$ neurons of the layer on each of the $\Nsamp$ samples remains the most time-consuming operation during the data preparation stage.

The proposed $\LayerDist$ metric, being a scalar value aggregating information on similarity at the layer level, inevitably leads to a loss of detailed information. Averaging distances over all optimally matched neuron pairs can mask the presence of specific neuron subgroups with significantly different or, conversely, very high similarity. In its current form, the metric does not account for the structural features of the matching itself (e.g., possible clustering of functionally similar neurons or the topology of their connections) and does not provide a mechanism for weighting the importance or contribution of individual neurons to the overall layer function.

There is also limited applicability for layers of different sizes. When comparing layers with different numbers of neurons ($N_1 \neq N_2$), the Hungarian algorithm finds matches for $\min(N_1, N_2)$ pairs. Neurons from the larger layer that remain unmatched are completely ignored when computing the current version of the $\LayerDist$ metric. This can lead to unintuitive or even paradoxical situations where, for example, adding new, functionally important neurons to one layer is not reflected (or is weakly reflected) in the final metric value. There is no natural mechanism for "penalizing" significant differences in the sizes of the compared layers.

Finally, it should be noted that generalizing the proposed static analysis to dynamic scenarios is complex. The method in its current form is intended for comparing "snapshots" of neural network states. Tracking the evolution of functional representations in a network over time (e.g., during training or adaptation) would require repeated full recalculation of costly neuron characteristics. There is no mechanism for incrementally updating signatures with small changes in network parameters, making it computationally very difficult to compare entire learning trajectories of different networks or to perform detailed analysis of dynamics at each iteration.

\bigskip
It is important to note that many of the listed limitations are not fundamental and can be viewed as directions for future research and improvements. The strengths of the method—its theoretical rigor, robustness to parameter variations, and potential for scalability—provide a solid foundation for further development. The limitations mainly indicate specific areas in which the proposed approach can be extended and improved, which is natural for pioneering research opening a new direction in neural network analysis.

\subsection{Defining promising directions for future work}
\label{subsec:future_work}

The approach proposed in this work and the developed similarity metric open significant prospects for further research and development. Several key directions can be highlighted for continuing this work.

One of the most relevant directions is extending the method to modern neural network architectures. Adapting it for convolutional neural networks (CNNs) will require developing the concept of an "activation region" for convolutional filters, taking into account their local receptive fields and translational invariance, as well as exploring ways to effectively represent spatial activation patterns and consider the hierarchical structure of features. For attention-based models, particularly transformers, fundamentally new approaches are needed to characterize attention heads through analysis of attention patterns on representative samples and the creation of specialized metrics to assess their functional similarity. Similarly, for recurrent architectures (RNN, LSTM, GRU), it will be necessary to develop methods to characterize their dynamic behavior and investigate temporal invariants in hidden states as a basis for constructing robust signatures.

The next important direction is the development of efficient hashing and compression methods for neural representations. This includes exploring alternative LSH schemes such as SimHash for continuous activation representations or hybrid approaches, as well as studying adaptive hashing methods (learnable hashing) that adjust to the task specifics. Improvements are also needed in adaptive tuning of hashing parameters, including automatic determination of the optimal number of hash functions $K$ and the creation of progressive schemes with incremental refinement capability.

Strategies for forming the sample $\XS$ also require further development and improvement. Promising work includes developing adaptive sampling methods, such as actively selecting the most informative points or using importance sampling. Special attention should be paid to handling high-dimensional data (images, texts), where it is necessary to explore dimensionality reduction methods and specialized sampling strategies for different modalities.

A substantial area of future work lies in theoretical research and formalization. Further development of the theory of canonical representations for various classes of neural networks is needed, along with the study of symmetry groups in modern architectures (beyond ReLU networks) and the development of algorithms for obtaining canonical forms with provable guarantees. Also of interest is analyzing the metric properties of the proposed $\LayerDist$ metric in limiting cases (e.g., as $N_k \to \infty$) and formalizing the conditions for metrizability of the neural layer space.

There is great potential in practical applications and the creation of corresponding tools. The proposed approach can be used to analyze the training process, including tracking the evolution of the network’s functional structure and identifying critical training phases. Based on it, methods for optimizing architectures can be developed, such as automatically searching for optimal network depth and width or pruning based on functional redundancy. Its application in federated learning for efficient and private model aggregation also deserves special attention.

Interdisciplinary directions are of interest as well. Investigating analogies between the proposed functional characteristics and the principles of organization of biological neural networks may lead to new discoveries in both fields. In the field of explainable artificial intelligence (XAI), functional signatures and similarity metrics can contribute to creating more interpretable representations of neuron behavior and explaining network decisions.

Finally, computational optimization and scaling of the proposed methods remain critically important directions. This includes developing optimized libraries for GPU-accelerated computation of functional signatures, designing distributed algorithms for analyzing very large models, and integrating with existing deep learning frameworks (PyTorch, TensorFlow, JAX). Special importance lies in developing incremental and online algorithms that allow updating signatures with small parameter changes without full recalculation.

These directions represent a natural evolution of the approach proposed in this work and open up broad prospects for both fundamental research in neural network theory and the creation of practical tools for analyzing and optimizing modern deep learning models.

\bibliographystyle{plain} 
\bibliography{reference} 

\end{document}